\documentclass[runningheads]{eccv2022submission}
\usepackage{graphicx}

\usepackage{tikz}
\usepackage{comment}
\usepackage{amsmath,amssymb} 
\usepackage{color}
\usepackage{booktabs}
\usepackage{mathtools}
\usepackage{xcolor,colortbl}
\usepackage{wrapfig}

\usepackage{bbm}
\usepackage{algorithm}
\usepackage{multirow}
\usepackage[noend]{algpseudocode}
\usepackage{pifont}%
\newcommand{\cmark}{\ding{51}}%
\newcommand{\xmark}{\ding{55}}%
\usepackage{subfig}

\usepackage[accsupp]{axessibility}  

\usepackage[pagebackref=true,breaklinks=true,colorlinks,bookmarks=false]{hyperref}

\hypersetup{citecolor=cyan, linkcolor=magenta}

\begin{document}
\pagestyle{headings}
\mainmatter
\def\ECCVSubNumber{7402}  



\title{Towards Realistic Semi-Supervised Learning} 

\titlerunning{Towards Realistic Semi-Supervised Learning}
%
\author{Mamshad Nayeem Rizve \and
Navid Kardan \and
Mubarak Shah}
\authorrunning{M. N. Rizve et al.}
%
\institute{Center for Research in Computer Vision, UCF, USA\\
\email{\{nayeemrizve, kardan\}@knights.ucf.edu, shah@crcv.ucf.edu}}
\maketitle

\begin{abstract}
Deep learning is pushing the state-of-the-art in many computer vision applications. However, it relies on large annotated data repositories, and capturing the unconstrained nature of the real-world data is yet to be solved. Semi-supervised learning (SSL) complements the annotated training data with a large corpus of unlabeled data to reduce annotation cost. The standard SSL approach assumes unlabeled data are from the same distribution as annotated data.  
Recently, a more realistic SSL problem, called open-world SSL, is introduced, where the unannotated data might contain samples from unknown classes. In this paper, we propose a novel pseudo-label based approach to tackle SSL in open-world setting. At the core of our method, we utilize sample uncertainty and incorporate prior knowledge about class distribution to generate reliable class-distribution-aware pseudo-labels for unlabeled data belonging to both known and unknown classes. Our extensive experimentation showcases the effectiveness of our approach on several benchmark datasets, where it substantially outperforms the existing state-of-the-art on seven diverse datasets including CIFAR-100 ($\sim$17\%), ImageNet-100 ($\sim$5\%), and Tiny ImageNet ($\sim$9\%). We also highlight the flexibility of our approach in solving novel class discovery task, demonstrate its stability in dealing with imbalanced data, and complement our approach with a technique to estimate the number of novel classes. Code: \url{https://github.com/nayeemrizve/TRSSL}
\end{abstract}

\section{Introduction}

Deep learning systems have made tremendous progress in solving many challenging vision problems \cite{he2016deep,he2017mask,chen2018encoder,girshick2015fast,Qi_2017_CVPR,Anderson_2018_CVPR}. However, most of this progress has been made in controlled environments, which limits their application in real-world scenarios. 
For instance, in classification, we should know all the classes in advance. However, many real-world problems cannot be expressed with this constraint, where we constantly encounter new concepts while exploring an unconstrained environment. A practical learning model should be able to properly detect and handle new situations. Open-world problems \cite{scheirer2012toward,bendale2015towards,kardan2016fitted,han2019learning,cao2022openworld,joseph2021towards,kardan2017mitigating} try to model this unconstrained nature of real-world data. 

Despite abundance of real-world data, it is often required to annotate raw data before passing it to supervised models, which is quiet costly. One of the dominant approaches to reduce the cost of annotation is semi-supervised learning (SSL) \cite{NIPS2017_6719_meanT,NIPS2019_8749_MixMatch,Lee2013PseudoLabelT,miyato2018virtual,sohn2020fixmatch}, where the objective is to leverage a set of unlabeled data in conjunction with a limited labeled set to improve performance. Following \cite{cao2022openworld}, in this work, we consider the unlabeled set to possibly contain samples from unknown (novel) classes that are not present in the labeled set. This problem is called open-world SSL \cite{cao2022openworld}. Here, the goal is to identify novel-class samples and classify them, as well as to improve known-class performance by utilizing unlabeled known-class samples.

At first sight, the major difficulty with open-world SSL might be related to breaking the closed-world assumption. In fact, it is common knowledge that presence of samples from novel classes deteriorates the performance of standard SSL methods drastically \cite{oliver2018realistic,chen2020semi}. This leads to introduction of new approaches that mitigate this issue based on identifying, and subsequently reducing the effect of novel class samples to generalize SSL to more practical settings \cite{guo2020safe,chen2020semi,zhao2020robust}. However, open-world SSL requires \emph{identifying and assigning samples to novel classes}, which contrasts with this simpler objective of ignoring them. To the best of our knowledge ORCA \cite{cao2022openworld} is the only prior work that proposes a solution for this challenging problem, where
the authors also demonstrate that open-world SSL problem cannot be solved by simple extensions of existing SSL approaches.

\begin{wrapfigure}{R}{0.49\textwidth}
\vspace{-8mm}
  \includegraphics[width=\linewidth]{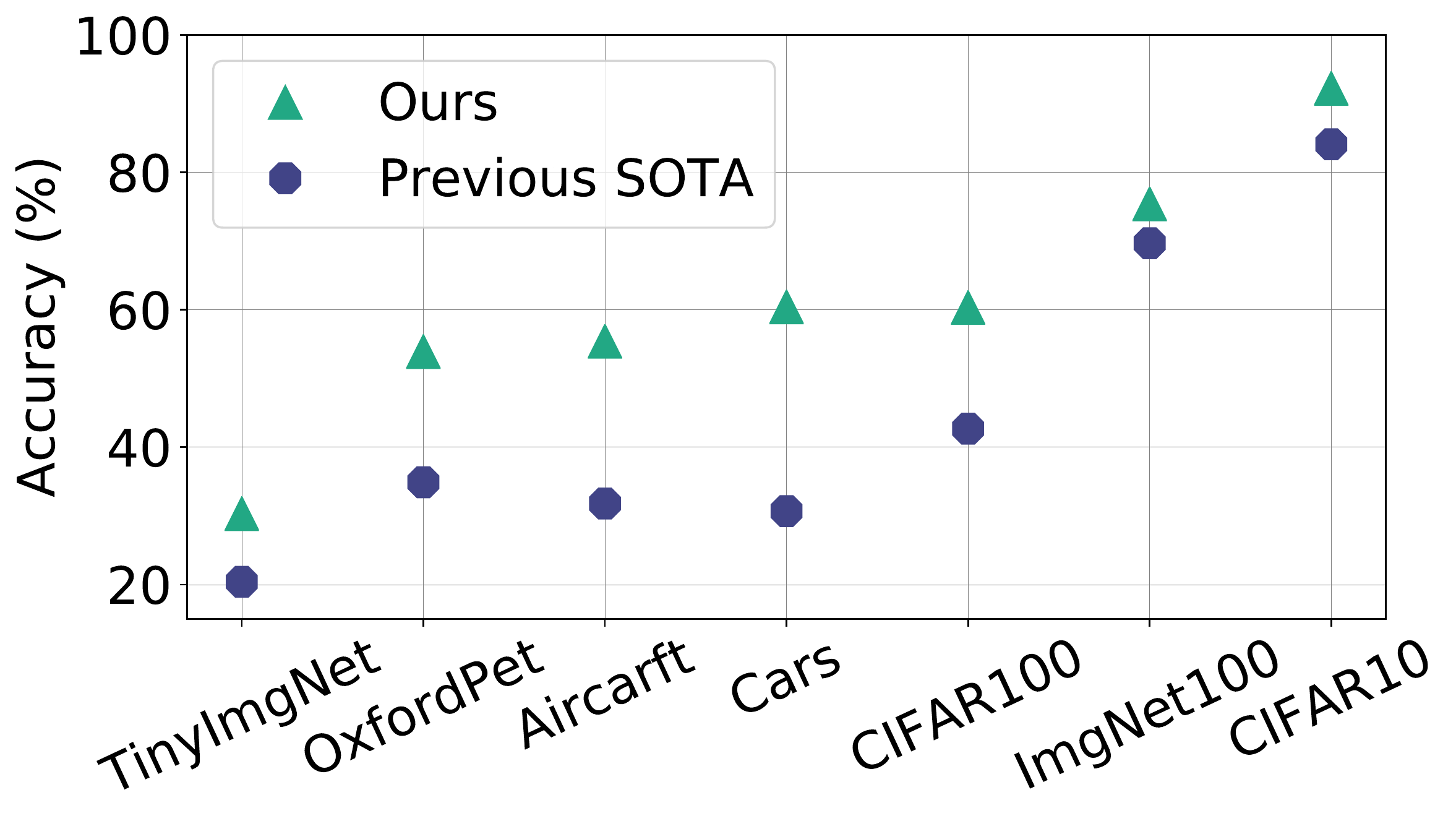}
\vspace{-8mm}
\caption{Performance of our proposed method with respect to previous SOTA method on {Tiny ImageNet}, {Oxford-IIIT Pet}, {FGVC-Aircraft}, {Stanford-Cars}, {CIFAR-100}, {ImageNet-100}, and {CIFAR-10} datasets respectively.}
\vspace{-8mm}
\label{fig:all_dataset}
\end{wrapfigure}

Improving upon ORCA, this paper introduces a streamlined approach for open-world SSL problem, which does not require careful design choices for multiple objectives, and does not rely on feature initialization. Our approach substantially improves state-of-the-art performance on multiple datasets (Fig.~\ref{fig:all_dataset}). Furthermore, distinctly from previous work, our algorithm can naturally handle arbitrary class distributions such as imbalanced data. Finally, we propose a method to estimate the number of unknown classes for more practical applications. 

For solving the open-world SSL problem, we employ an intuitive pseudo-labeling approach. Our pseudo-label generation process takes different challenges associated with the open-world SSL problem---simultaneously classifying samples from both known and unknown classes, and handling arbitrary class distribution---into account. Furthermore, we incorporate sample uncertainty into pseudo-label learning to address the unreliable nature of generated pseudo-labels. We make two major technical contributions in this work: (1) we propose a novel pseudo-label generation method, which takes advantage of the prior knowledge about class distribution and generate pseudo-labels accordingly using Sinkhorn-Knopp algorithm \cite{sinkhorn1967concerning,caron2020unsupervised,YM.2020Self-labelling,asano2020labelling}. Our proposed solution can take advantage of any arbitrary data distribution which includes imbalanced distributions. (2) we introduce a novel uncertainty-guided temperature scaling technique to address the unreliable nature of the generated pseudo-labels. Additionally, we propose a simple yet effective method for estimating the number of novel classes, allowing for a more realistic application of our method. Our extensive experimentation on four standard benchmark datasets and also three additional fine-grained datasets demonstrate that the proposed method significantly outperforms the existing works (Fig.~\ref{fig:all_dataset}). Finally, our experimentation with data imbalance (Sec.~\ref{par:data_imbalance}) signifies that the proposed method can work satisfactorily even when no prior knowledge is available about the underlying class distribution.

\section{Related Works}
\textbf{Open-World Learning}
To address the unconstrained nature of real-world data, multiple research directions have been explored. In this work, we refer to all these different approaches as open-world learning method. Open-set recognition (OSR) \cite{scheirer2012toward,jain2014multi,liang2017enhancing}, open-world recognition (OWR) \cite{bendale2015towards,boult2019learning,xu2019open,joseph2021towards}, out-of-distribution detection \cite{liang2017enhancing,lee2018simple,zaeemzadeh2021out,kardan2021towards,kardan2021self}, and novel class discovery (NCD) \cite{hsu2018learning,han2019learning,Han2020Automatically,Zhong_2021_CVPR,fini2021unified} are some of the notable open-world learning approaches.

Open-set recognition methods aim to identify novel class samples during inference to avoid assigning them to one of the known/seen classes. 
One of the early works on OSR was proposed in \cite{scheirer2012toward}, where a one-vs-all strategy was applied to prevent assigning novel class samples to known classes. \cite{jain2014multi} extends OSR to multi-class setup by using probabilistic modeling to adjust the classification boundary. Instead of designing robust models for OSR, ODIN \cite{liang2017enhancing} detects novel class samples (out-of-distribution) based on difference in output probabilities caused by changing the softmax temperature and adding small controlled perturbations to the inputs. Even though OSR is a related problem, the focus of this work is more general where our goal is to not only detect novel class samples but also to cluster them.

OWR methods such as \cite{bendale2015towards} work in an incremental manner, where once the model determines instances from novel classes an oracle can provide class labels for unknown samples to incorporate them into the seen set. To incorporate new classes, \cite{xu2019open} maintains a dynamic list of exemplar samples for each class, and unknown examples are detected by finding the similarity with these exemplars. Finally, authors in \cite{joseph2021towards} propose contrastive clustering and energy based unknown sample detection for open-world object detection. The key difference between these methods and ours is that we do not rely on an oracle to learn novel classes.       

NCD methods are most closely related to our task. The main objective of NCD methods is to cluster novel class samples in the unlabeled set. To this end, authors in \cite{hsu2018learning} leverage the information available in the seen classes by training a pairwise similarity prediction network that they later apply to cluster novel class samples. Similar to their approach, a pairwise similarity task is solved  to discover novel classes based on a novel rank statistics in \cite{Han2020Automatically}. Most NCD methods rely on multiple objective functions and require some sort of feature pretraining approach. This is addressed in \cite{fini2021unified}  by utilizing multi-view pseudo-labeling and overclustering while only relying on cross-entropy loss. The main difference between NCD problem and our task is that we do not assume unlabeled data only includes novel class samples. Besides, in contrast to most of these methods, our proposed solution requires only one loss function and does not make architectural changes to treat seen and novel classes differently. Additionally, our extensive experimentation demonstrates that extension of these methods is not very effective for open-world SSL problem.

\vspace{1mm}
\noindent\textbf{Semi-Supervised Learning } 
Extensive research has been conducted on closed-world SSL \cite{Gammerman1998Learning,joachims1999transductive,liu2019deep,kingma2014semi,pu2016variational,chen2020big,caron2020unsupervised,NIPS2016_6333,LaineA17,Miyato2018VirtualAT,NIPS2017_6719_meanT,Lee2013PseudoLabelT,Shi_2018_ECCV,NIPS2019_8749_MixMatch,Berthelot2020ReMixMatch:,sohn2020fixmatch}. The closed-world SSL methods achieve impressive performance on standard benchmark datasets. However, these methods assume that the unlabeled data only contains samples from seen classes, which is very restrictive. Moreover, recent works \cite{oliver2018realistic,chen2020semi} suggest that presence of novel class samples deteriorates performance of SSL methods. Robust SSL methods \cite{guo2020safe,chen2020semi,zhao2020robust} address this issue by filtering out or reweighting novel class samples. The realistic open-world SSL problem as proposed in \cite{cao2022openworld} requires clustering the novel class samples which is not addressed by robust SSL methods. To the best of our knowledge, ORCA \cite{cao2022openworld} is the only prior work that solves this challenging problem. ORCA achieves very promising performance in comparison to other novel class discovery or robust SSL based baselines. However, to solve this problem ORCA leverages self-supervised pretraining and multiple objective functions. In a concurrent work, \cite{rizve2022openldn} proposes a solution to open-world SSL that does not rely on feature pretraining. However, similar to ORCA, their approach relies on multiple objectives. In contrast, our proposed solution outperforms ORCA by a large margin without relying on either of them.

\vspace{-2mm}
\section{Method}
\vspace{-2mm}

Similar to standard closed-world SSL, the training data for open-world SSL problem consists of a labeled set, $\mathbb{D}_L$, and an unlabeled set, $\mathbb{D}_U$. The labeled set, $\mathbb{D}_L$ encompasses $N_L$ labeled samples s.t.\ $\mathbb{D}_L = \big\{\mathbf{x}^{(i)}_l, {\mathbf{y}}^{(i)}_l\big\}_{i=1}^{N_L}$, where $\mathbf{x}^{(i)}_l$ is an input and ${\mathbf{y}}^{(i)}_l$ is its corresponding label (in one-hot encoding) belonging to one of the $\mathbb{C}_L$ classes. On the other hand, the unlabeled set, $\mathbb{D}_U$, consists of $N_U$ (\text{in practice, } $N_U\gg N_L$) unlabeled samples s.t.\ $\mathbb{D}_U=\big\{\mathbf{x}^{(i)}_u\big\}_{i=1}^{N_U},$ where $\mathbf{x}^{(i)}_u$ is a sample without any label that belongs to one of the $\mathbb{C}_U$ classes. The primary distinction between the closed-world and open-world SSL formulation is that the closed-world SSL assumes $\mathbb{C}_L = \mathbb{C}_U$, whereas in open-world SSL $\mathbb{C}_L \subset \mathbb{C}_U$. We refer to $\mathbb{C}_U\setminus \mathbb{C}_L$, as novel classes, $\mathbb{C}_N$. Note that unlike previous works on novel class discovery problem \cite{Han2020Automatically,fini2021unified,Zhong_2021_CVPR}, we do not need to know the number of novel classes, $|\mathbb{C}_N|$, in advance. During test time, the objective is to assign samples from novel classes to their corresponding novel class in $\mathbb{C}_N$, and to classify the samples from seen classes into one of the $|\mathbb{C}_L|$ classes.

In the following subsections, we first introduce our class-distribution-aware pseudo-label based training objective to classify the samples from seen classes, while attributing the samples from novel classes to their respective categories (Sec.~\ref{sec:equi-pl}). After that, we introduce uncertainty-guided temperature scaling to incorporate reliability of pseudo-labels into the learning process (Sec.~\ref{sec:uncr-temp}).

\begin{figure*}[h]
\vspace{2mm}
\begin{center}
  \includegraphics[width=1.0\linewidth]{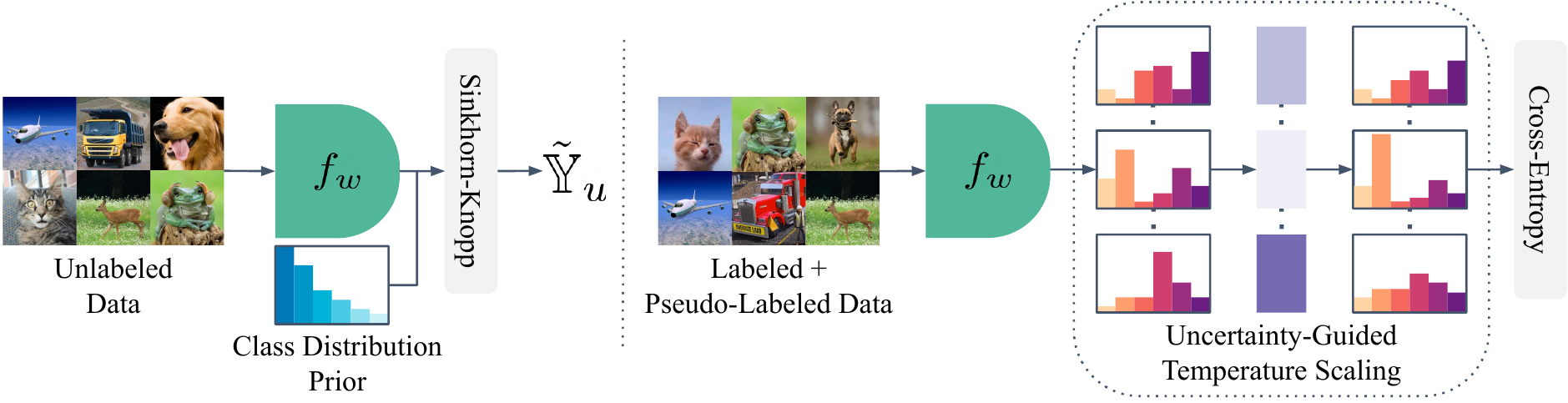}
\end{center}
\vspace{-4mm}
\caption{\textit{Training Overview:} \textbf{Left:} generating pseudo-labels. Our model generates pseudo-labels for the unlabeled samples using Sinkhorn-Knopp while taking class distribution prior into account. \textbf{Right:} reliable training with both labeled and unlabeled samples. We use the ground-truth labels and generated pseudo-labels to train in a supervised manner
. To address the unreliable nature of pseudo-labels in open-world SSL, we apply uncertainty-guided temperature scaling (darker color refers to higher uncertainty).}
\vspace{-6mm}

\label{fig:arch}
\end{figure*}

\subsection{Class-Distribution-Aware Pseudo-Labeling}
\label{sec:equi-pl}

To achieve the dual objective of open-world SSL problem, i.e., identifying samples from the seen classes and clustering the samples from novel classes, we design a single classification objective. To this end, we utilize a neural network, $f_w$, to map the input data $\mathbf{x}$ into the output space of class scores (logits), $\mathbf{{z}}\in \mathbb{R}^{|\mathbb{C}_L| + |\mathbb{C}_N|}$, s.t.\ $f_w: \mathbb{X} \rightarrow \mathbb{{Z}}$; here, $\mathbb{X}$ is the set of input data and $\mathbb{Z}$ is the set of output logits. In our setup, the first $|\mathbb{C}_L|$ entries of the class score vector (logits), $\mathbf{z}$, correspond to seen classes and the remaining $|\mathbb{C}_N|$ elements correspond to novel classes. Finally, we transform these logits to probability distribution, $\mathbf{\Hat{y}}$, using softmax activation function: $\mathbf{\hat{y}}_j = \mathrm{exp}(\mathbf{z_j})/\sum_k\mathrm{exp}(\mathbf{z_k})$.

The neural network, $f_w$, can be trained using cross-entropy loss if the labels for all the input samples are available. However, in open-world SSL problem the samples in $\mathbb{D}_U$ lack label. To address this issue, pseudo-labels, $\mathbf{\Tilde{y}}_u\in\mathbb{\Tilde{Y}}_u$, are generated for all unlabeled samples. After that, cross entropy loss is  applied to train the model using the available ground-truth labels, ${\mathbf{y}}_l\in{\mathbb{Y}}_l$, and generated pseudo-labels. Here, we assume one-hot encoding for ${\mathbf{y}}_l$ and ${\mathbb{Y}}$ denotes the set of all labels, where ${\mathbb{Y}} = \mathbb{Y}_l\cup \mathbb{\Tilde{Y}}_u$. Now, the cross-entropy loss is defined using,

\setlength{\abovedisplayskip}{-6pt}
\setlength{\belowdisplayskip}{2pt}
\setlength{\abovedisplayshortskip}{0pt}
\setlength{\belowdisplayshortskip}{0pt}

\begin{align}
\label{eqn:ce}
   \mathcal{L}_{ce} = {-\frac{1}{N}\sum_{i=1}^{N}\sum_{j=1}^{C}\mathbf{y}^{(i)}_j}\log\mathbf{\hat{y}}^{(i)}_j, 
\end{align}
where, $C={|\mathbb{C}_L| + |\mathbb{C}_N|}$ is total number of classes, $N={N_L + N_U}$ is the total number of samples, $\mathbf{y}\in\mathbb{Y}$, and $\mathbf{y}^{(i)}_j$ is the $j$th element of the class label vector, $\mathbf{y}^{(i)}$, for training instance $i$.  

Next, we discuss the class-distribution-aware pseudo-label generation process. Since pseudo-label generation process is inherently ill-posed, we can guide this process by injecting an inductive bias. To this end, we propose to generate pseudo-labels in such a way that the class distribution of generated pseudo-labels should follow the underlying class distribution of samples. More formally, we enforce the following constraint: 

\setlength{\abovedisplayskip}{-6pt}
\setlength{\belowdisplayskip}{2pt}
\setlength{\abovedisplayshortskip}{0pt}
\setlength{\belowdisplayshortskip}{0pt}

\begin{align}
\label{eqn:eqpartition}
   \forall j\sum_i^{N_U}\mathbf{\tilde{y}}_{j}^{(i)} = N_{U}^{C_j}, 
\end{align}

\noindent where, $N_{U}^{C_j}$ is the number of samples in $j$th class.

One common strategy to satisfy this objective is to apply an entropy maximization term coupled with optimizing a pairwise similarity score objective \cite{cao2022openworld,van2020scan}. However, this approach implicitly assumes that the classes are balanced and optimizing the pairwise objective requires a good set of initial features; besides, coordinating these two objectives requires careful design. This paper pursues a more streamlined approach by generating pseudo-labels such that they directly satisfy the constraints in Eq.~\ref{eqn:eqpartition}. Fortunately, this constrained pseudo-label generation problem is inherently a transportation problem \cite{kantorovich1942translation,brenier1987decomposition}, where we want to assign unlabeled samples to one of the seen/novel classes based on output probabilities. Such an assignment can be captured with an assignment matrix, $\mathbf{A}$, which can be interpreted as (normalized) pseudo-labels. Following Cuturi's notation \cite{cuturi2013sinkhorn}, every such assignment $\mathbf{A}$, called a transport matrix, that satisfies the constraint in Eq.~\ref{eqn:eqpartition} is a member of a transportation polytope, $\mathcal{A}$. 

\setlength{\abovedisplayskip}{-6pt}
\setlength{\belowdisplayskip}{2pt}
\setlength{\abovedisplayshortskip}{0pt}
\setlength{\belowdisplayshortskip}{0pt}

\begin{align}
   \mathcal{A} := \Big\{\mathbf{A}\in \mathbb{R}^{N_U\times C}|
   \forall j \sum \mathbf{A}_{:,j}=\frac{N_{U}^{C_j}}{N_U},
   \forall i \sum \mathbf{A}_{i,:}=\frac{1}{N_U}\Big\}.
\end{align}

Note that every transport matrix $\mathbf{A}$ is a joint probability, therefore, it is a normalized matrix. By considering the cross-entropy cost of assigning unlabeled samples based on model predictions to different classes, an optimal solution can be found within the transportation polytope $\mathcal{A}$. More formally, we solve $\min_{\mathbf{A}\in\mathcal{A}}-Tr(\mathbf{A}^T\log(\mathbf{\hat{Y}}_U/N_U))$ optimization problem, where $\mathbf{\hat{Y}}_U$ is the matrix of output probabilities generated by the model for the unlabeled samples. Unfortunately, enforcing the constraint described in Eq.~\ref{eqn:eqpartition} is non-trivial for novel classes since we do not know the specific order of novel classes. To address this issue, we need to solve a permutation problem while obtaining the optimal assignment matrix, $\mathbf{A}$. To this end, we introduce a permutation matrix $\mathbf{P_\pi}$ and reformulate the optimization problem as $\min_{\mathbf{A}\in\mathcal{A}}-Tr((\mathbf{A}\mathbf{P_\pi})^T\log(\mathbf{\hat{Y}}_U/N_U))$. Here, the permutation matrix $\mathbf{P_\pi}$ reorders the columns of the assignment matrix. We estimate the permutation matrix $\mathbf{P_\pi}$ from the order of the marginal of output probabilities $\mathbf{\hat{Y}}_U$. This simple reordering ensures that per class constraint is aligned with the output probabilities. After determining the permutation, finding the optimal solution for $\mathbf{A}$ becomes an instance of the optimal transport problem. Hence, can be solved using Sinkhorn-Knopp algorithm. Cuturi \cite{cuturi2013sinkhorn} proposes a fast version of Sinkhorn-Knopp algorithm. In particular, \cite{cuturi2013sinkhorn} shows that a fast estimation of the optimal assignment can be obtained by:

\setlength{\abovedisplayskip}{-10pt}
\setlength{\belowdisplayskip}{2pt}
\setlength{\abovedisplayshortskip}{0pt}
\setlength{\belowdisplayshortskip}{0pt}

\begin{align}
\label{eqn:sinkhon1}
   \mathbf{A} = \mathrm{diag}(\mathbf{m})(\mathbf{\Hat{Y}}_U/N_U)^{\lambda}\mathrm{diag}(\mathbf{n}),
\end{align}

\noindent where $\lambda$ is a regularization term that controls the speed of convergence versus precision of the solution, vectors $\mathbf{m}$ and $\mathbf{n}$ are used for scaling $\mathbf{\Hat{Y}}_U/N_U$ so that the transportation matrix $\mathbf{A}$ is also a probability matrix. This is an itereative procedure where $\mathbf{m}$ and $\mathbf{n}$ are updated according to the following rules:

\setlength{\abovedisplayskip}{-6pt}
\setlength{\belowdisplayskip}{2pt}
\setlength{\abovedisplayshortskip}{0pt}
\setlength{\belowdisplayshortskip}{0pt}

\begin{align}
\label{eqn:sinkhon2}
   \mathbf{m} \gets [(\mathbf{\Hat{Y}}_U/N_U)^{\lambda}\mathbf{n}]^{-1}, \mathbf{n} \gets [\mathbf{m}^T(\mathbf{\Hat{Y}}_U/N_U)^{\lambda}]^{-1}.
\end{align}

Another aspect of our pseudo-label generation is inducing perturbation invariant features. Generally learning invariant features is achieved by minimizing a consistency loss that minimizes the distance 
between the output representation of two transformed versions of the same image \cite{NIPS2016_6333,NIPS2019_8749_MixMatch,Verma2019InterpolationCT}. To achieve this, for the unlabeled data, given image $\mathbf{x}$, we generate two augmented versions of this image, $\mathbf{x}_{\tau_1}=\tau_1(\mathbf{x})$, and $\mathbf{x}_{\tau_2}=\tau_2(\mathbf{x})$, where $\tau_1(.)$, and $\tau_2(.)$ are two stochastic transformations. The generated pseudo-labels for these two augmented images are $\mathbf{\tilde{y}}_{\tau_1}$, and $\mathbf{\tilde{y}}_{\tau_2}$, respectively. To learn transformation invariant representation using cross-entropy loss, we treat $\mathbf{\tilde{y}}_{\tau_2}$ as the corresponding pseudo-label of $\mathbf{x}_{\tau_1}$ and vice versa. This cross pseudo-labeling encourages learning of perturbation invariant features without introducing a new loss function.

Finally, in its original formulation Sinkhorn-Knopp algorithm generates hard pseudo-labels \cite{cuturi2013sinkhorn}. However, recent literature \cite{caron2020unsupervised} reports better performance by applying soft pseudo-labels. In our work we utilize a mixture of soft and hard pseudo-labels (mixed pseudo-labels), which we found to be beneficial (Sec.~\ref{sec:ablation_analysis}). To be specific, to encourage confident learning for novel classes, we generate hard pseudo-labels for unlabeled samples which are strongly assigned to novel classes. For the rest of the unlabeled samples, we use soft pseudo-labels.

\subsection{Uncertainty-Guided Temperature Scaling}
\label{sec:uncr-temp}
Since we generate pseudo-labels by relying on the confidence scores of the network, final performance is affected by their reliability. We can capture the reliability of prediction confidences by measuring their uncertainty. One simple way to do that in the standard neural networks is to perform Monte Carlo sampling in the network parameter space \cite{gal2016dropout} or in the input space \cite{ayhan2018,rizve2021in}. Since we do not want to modify the network parameters, we decide to perform stochastic sampling in input space. To this end, we apply stochastic transformations on input data  and estimate the sample uncertainty, $\mathrm{u}(.)$, by calculating the variance over the applied stochastic transformations \cite{feinman2017detecting,NEURIPS2020_f23d125d,rizve2021in}:

\setlength{\abovedisplayskip}{-6pt}
\setlength{\belowdisplayskip}{2pt}
\setlength{\abovedisplayshortskip}{0pt}
\setlength{\belowdisplayshortskip}{0pt}

\begin{align}
\label{eqn:var}
    \mathrm{u}({\mathbf{x}})=\mathrm{Var}(\mathbf{\hat{y}}) = \frac{1}{\mathcal{T}}\sum_{i=1}^\mathcal{T}(\mathbf{\hat{y}}_{\tau_i}-\mathrm{E}(\mathbf{\hat{y}}))^2,
\end{align}

\noindent where, $\mathbf{\hat{y}}_{\tau_i} = \mathrm{Softmax}(f_w(\tau_i(\mathbf{x})))$, $\tau_i(.)$ represents a stochastic transformation applied to the input $\mathbf{x}$, and $\mathrm{E}(\mathbf{\hat{y}})=\frac{1}{\mathcal{T}}\sum_{i=1}^\mathcal{T}\mathbf{\hat{y}}_{\tau_i}$.

Next, we want to incorporate this uncertainty information into our training process. One strategy to achieve this is to select more reliable pseudo-labels by filtering out unreliable samples based on their uncertainty score \cite{rizve2021in}. However, two potential drawbacks of this approach are introducing a new hyperparameter and discarding a portion of available data. Therefore, to tackle both of these drawbacks, we introduce uncertainty-guided temperature scaling.   

Recall that in our training we use softmax probabilities for cross-entropy loss. Temperature scaling is a strategy to modify the softness of the output probability distribution. In standard softmax probability computation, the temperature value is set to 1. A higher value of temperature increases the entropy or uncertainty of the softmax probability, whereas a lower value makes it more certain. Existing works \cite{cao2022openworld,fini2021unified,chen2020simple,khosla2020supervised} apply a fixed temperature value (whether high or low) as a hyperparameter. In contrast, we propose to use a different temperature for each sample during the course of training which is influenced by the certainty of its pseudo-label. The main idea is that if the network is certain about its prediction on a particular sample we make this prediction more confident and vice versa. Based on this idea we modify the softmax probability computation in the following way:

\setlength{\abovedisplayskip}{-6pt}
\setlength{\belowdisplayskip}{2pt}
\setlength{\abovedisplayshortskip}{0pt}
\setlength{\belowdisplayshortskip}{0pt}
\begin{align}
\label{eqn:uts}
    \mathbf{\hat{y}}^{(i)}_j = \frac{ \mathrm{exp}(\mathbf{z}^{(i)}_j/\mathrm{u}(\mathbf{x}^{(i)}))}{\sum_k\mathrm{exp}(\mathbf{z}^{(i)}_k/\mathrm{u}(\mathbf{x}^{(i)}))},
\end{align}
where $\mathrm{u}(\mathbf{x}^{(i)})$ is the uncertainty of sample $\mathbf{x}^{(i)}$ that is obtained from Eq.~\ref{eqn:var}.

In practice, the sample uncertainties calculated by Eq.~\ref{eqn:var} have low magnitudes. Therefore, we normalize these uncertainty values across the entire dataset before plugging them into Eq.~\ref{eqn:uts}.

Our training algorithm is provided in supplementary materials. 

\section{Experiments and Results}
\label{sec:exp}
\subsection{Experimental Setup}
In the following, we describe our experimental setup including dataset descriptions, implementation details, evaluation details, and specifics of our baselines. 

\vspace{1mm}
\label{para:dataset}
\noindent \textbf{Datasets} We conduct experiments on four commonly used computer vision benchmark datasets: CIFAR-10 \cite{cifar10}, CIFAR-100 \cite{cifar100}, ImageNet-100 \cite{russakovsky2015imagenet} and Tiny ImageNet \cite{le2015tiny}. The datasets are selected in increasing order of difficulty based on the number of classes. We also evaluate our method on three drastically different fine-grained classification datasets: Oxford-IIIT Pet \cite{parkhi12a}, FGVC-Aircraft \cite{maji13fine-grained}, and Stanford-Cars \cite{KrauseStarkDengFei-Fei_3DRR2013}. A detailed description of these datasets is provided in supplementary materials. For all the datasets, we use the first 50\% classes as seen and the remaining 50\% classes as novel. We use 10\% data from the seen classes as the labeled set and use the remaining 90\% data, along with the samples from novel classes, as unlabeled set for our experiments on standard benchmark datasets. For fine-grained datasets, we use 50\% data from seen classes as labeled. Additional results with other data percentage are provided in the supplementary materials.

\vspace{1mm}
\label{para:implimentation}
\noindent \textbf{Implementation Details} Following ORCA \cite{cao2022openworld}, for a fair comparison, we use ResNet-50 \cite{he2016deep} for ImageNet-100 experiments and ResNet-18 \cite{he2016deep} for all the other experiments. We apply $l_2$ normalization to the weights of the last linear layer. For CIFAR-10, CIFAR-100, and Tiny ImageNet experiments, we train our model for 200 epochs. For the other datasets, we train these model for 100 epochs. We use a batchsize of 256 for all of our experiments except ImageNet-100 where similar to \cite{cao2022openworld} we use a batchsize of 512. For optimizing the network parameters, we use SGD optimizer with momentum. We use a cosine annealing based learning rate scheduler accompanied by a linear warmup, where we set the base learning rate to 0.1 and set the warmup length to 10 epochs. For network parameters, we set the weight decay to 1e-4. Following \cite{caron2020unsupervised}, we set the value of $\lambda$ to 0.05 and perform 3 iterations for pseudo-label generation using the Sinkhorn-Knopp algorithm. Additional implementation details are provided in supplementary materials.

\vspace{1mm}
\noindent \textbf{Evaluation Details} For evaluation, we report standard classification accuracy on seen classes. On novel classes, we report clustering accuracy following \cite{cao2022openworld,Han2020Automatically,fini2021unified,han2019learning}. To this end, we consider the class prediction as cluster ID. Next, we use the Hungarian algorithm \cite{kuhn1955hungarian} to match cluster IDs with ground-truth classes. Once the matches are obtained, we calculate classification accuracy with the corresponding cluster IDs. Besides, if a novel class sample gets assigned to one of the seen classes, we consider that as a misclassified prediction and remove that sample before matching the cluster IDs with ground-truth class labels. We also report clustering accuracy for all the classes.

\vspace{1mm}
\noindent \textbf{Comparison Details} We compare the performance of our method on CIFAR-10, CIFAR-100, and ImageNet-100 datasets with the results reported in \cite{cao2022openworld}. The remaining four datasets do not have any publicly available evaluation for open-world SSL problem. Therefore, we extend three recent novel class discovery methods \cite{Han2020Automatically,han2019learning,fini2021unified} to open-world SSL setting using publicly available codebase. For \cite{Han2020Automatically,han2019learning}, we extend the unlabeled head to include logits for seen classes by following \cite{cao2022openworld}. However, neither of these methods has any explicit classification loss for seen classes in the unlabeled head. Therefore, there is no straightforward way to map the seen class samples into their corresponding class logits. For reporting scores on seen classes, we use the Hungarian algorithm for these two methods. In \cite{fini2021unified}, pseudo-labels are generated for the novel class samples on the unlabeled head. To make it compatible with open-world SSL setting, we generate pseudo-labels from the concatenated prediction of the labeled and unlabeled heads during training. Since this method has explicit classification loss, we report standard classification accuracy on seen classes.

\subsection{Main Results}
\noindent \textbf{Standard Benchmark Datasets}
We compare our method with existing literature on open-world SSL problem \cite{cao2022openworld} and other related approaches that have been modified for this problem in Tab.~\ref{tab:vision} and \ref{tab:finegrained}. On CIFAR-10 we observe that our proposed method outperforms ORCA \cite{cao2022openworld} on both seen and novel classes by 12.1\% and 4.1\%, respectively. Our proposed method also outperforms other novel class discovery methods \cite{han2019learning,Han2020Automatically,fini2021unified} by a large margin. The same trend is observed for FixMatch \cite{sohn2020fixmatch} (a state-of-the-art closed-world SSL method). Finally, our proposed method outperforms DS$^3$L\cite{guo2020safe}, a popular robust SSL method. Interestingly, improvement of our proposed method is more prominent on CIFAR-100 dataset, which is more challenging because of the higher number of classes. On CIFAR-100 dataset, our proposed method outperforms ORCA by around 20\% on novel classes and 16\% on seen classes. Noticeably, we observe that UNO\cite{fini2021unified} marginally outperforms ORCA on this dataset. However, our proposed method outperforms UNO by a significant margin. Next, we evaluate on two variants of ImageNet: ImageNet-100, and Tiny ImageNet. We observe a similar trend on ImageNet-100 dataset, where we observe an overall improvement of 5.7\% over ORCA. After that, we conduct experiments on challenging Tiny ImageNet dataset. This dataset is more challenging than CIFAR-100 and ImageNet-100 dataset since it has 200 classes. Besides, without transfer learning, even the performance of supervised methods is relatively low on this dataset. Overall, our proposed method outperforms the second best method, UNO, by 9.9\%, which is almost 50\% relative improvement on this challenging dataset. The results on these four datasets demonstrate that \emph{the proposed method not only outperforms previous methods but also excels in scenarios where the number of classes is significantly higher which is always a challenge for clustering methods.}

\begin{table*}[t]
\caption{Average accuracy on the \textbf{CIFAR-10}, \textbf{CIFAR-100}, and \textbf{ImageNet-100} datasets with 50\% classes as seen and 50\% classes as novel. The results are averaged over three independent runs.}
\begin{center}\setlength{\tabcolsep}{2pt}
\small
\resizebox{0.83\textwidth}{!}{%
\begin{tabular}{lccc|ccc|ccc}
\hline


 \multicolumn{1}{l}{\multirow{2}{*}{\textbf{Method}}} &
 \multicolumn{3}{c|}{\textbf{CIFAR-10}} & \multicolumn{3}{c|}{\textbf{CIFAR-100}} & \multicolumn{3}{c}{\textbf{ImageNet-100}} \\  
\multicolumn{1}{c}{} & \textbf{Seen} & \textbf{Novel} & \textbf{All} & \textbf{Seen} & \textbf{Novel} & \textbf{All}  & \textbf{Seen} & \textbf{Novel} & \textbf{All}\\


\hline
FixMatch\cite{sohn2020fixmatch} & $64.3$ & $49.4$ & $47.3$ & $30.9$ & $18.5$ & $15.3$ & $60.9$ & $33.7$ & $30.2$ \\
DS$^{3}$L\cite{guo2020safe} & $70.5$ & $46.6$ & $43.5$ & $33.7$ & $15.8$ & $15.1$ & $64.3$ & $28.1$ & $25.9$\\
DTC\cite{han2019learning} & $42.7$ & $31.8$ & $32.4$ & $22.1$ & $10.5$ & $13.7$ & $24.5$ & $17.8$ & $19.3$\\
RankStats\cite{Han2020Automatically} & $71.4$ & $63.9$ & $66.7$ & $20.4$ & $16.7$ & $17.8$ & $41.2$ & $26.8$ & $37.4$\\
UNO\cite{fini2021unified} & $86.5$ & $71.2$ & $78.9$ & $53.7$ & $33.6$ & $42.7$ & $66.0$ & $42.2$ & $53.3$\\
ORCA\cite{cao2022openworld} & $82.8$ & $85.5$ & $84.1$ & $52.5$ & $31.8$ & $38.6$ & {\cellcolor{yellow!15}}${83.9}$ & $60.5$ & $69.7$\\
Ours & {\cellcolor{yellow!15}}${94.9}$ & {\cellcolor{yellow!15}}${89.6}$ & {\cellcolor{yellow!15}}${92.2}$ & {\cellcolor{yellow!15}}${68.5}$ & {\cellcolor{yellow!15}}${52.1}$ & {\cellcolor{yellow!15}}${60.3}$ & $82.6$ & {\cellcolor{yellow!15}}${67.8}$ & {\cellcolor{yellow!15}}${75.4}$\\ 
\hline 


\end{tabular}
}
\end{center}

\label{tab:vision}
\vspace{-6mm}
\end{table*}

\begin{table*}[t]
\caption{Average accuracy on the \textbf{Tiny ImageNet}, \textbf{Oxford-IIIT Pet}, \textbf{FGVC-Aircraft}, and \textbf{Stanford-Cars}  datasets with 50\% classes as seen and 50\% classes as novel. The results are averaged over three independent runs.}
\begin{center}\setlength{\tabcolsep}{2pt}
\small
\resizebox{\textwidth}{!}{%
\begin{tabular}{lccc|ccc|ccc|ccc}
\hline


 \multicolumn{1}{l}{\multirow{2}{*}{\textbf{Method}}} &
 \multicolumn{3}{c|}{\textbf{Tiny ImageNet}} & \multicolumn{3}{c|}{\textbf{Oxford-IIIT Pet}} & \multicolumn{3}{c|}{\textbf{FGVC-Aircraft}} &\multicolumn{3}{c}{\textbf{Stanford-Cars}} \\  
\multicolumn{1}{c}{} & \textbf{Seen} & \textbf{Novel} & \textbf{All} & \textbf{Seen} & \textbf{Novel} & \textbf{All}  & \textbf{Seen} & \textbf{Novel} & \textbf{All}  & \textbf{Seen} & \textbf{Novel} & \textbf{All}\\


\hline
DTC\cite{han2019learning} & $13.5$ & $12.7$ & $11.5$ & $20.7$ & $16.0$ & $13.5$ & $16.3$ & $16.5$ & $11.8$ & $12.3$ & $10.0$ & $7.7$\\
RankStats\cite{Han2020Automatically} & $9.6$ & $8.9$ & $6.4$ & $12.6$ & $11.9$ & $11.1$ & $13.4$ & $13.6$ & $11.1$ & $10.4$ & $9.1$ & $6.6$ \\
UNO\cite{fini2021unified} & $28.4$ & $14.4$ & $20.4$ & $49.8$ & $22.7$ & $34.9$ & $44.4$ & $24.7$ & $31.8$ & $49.0$ & $15.7$ & $30.7$\\
Ours & {\cellcolor{yellow!15}}${39.5}$ & {\cellcolor{yellow!15}}${20.5}$ & {\cellcolor{yellow!15}}${30.3}$ & {\cellcolor{yellow!15}}${70.9}$ & {\cellcolor{yellow!15}}${36.1}$ & {\cellcolor{yellow!15}}${53.9}$ & {\cellcolor{yellow!15}}${69.5}$ & {\cellcolor{yellow!15}}${41.2}$ & {\cellcolor{yellow!15}}${55.4}$ & {\cellcolor{yellow!15}}${83.5}$ & {\cellcolor{yellow!15}}${37.1}$ & {\cellcolor{yellow!15}}${60.4}$\\\hline 


\end{tabular}
}
\end{center}
\label{tab:finegrained}
\vspace{-6mm}
\end{table*}

\vspace{1mm}
\noindent \textbf{Fine-Grained Datasets} Finally, we evaluate our method on three fine-grained classification datasets with different number of classes. This evaluation is particularly important since fine-grained classification captures challenges associated with many real-world applications. We hypothesize that, fine-grained classification is a harder problem for open-world semi-supervised learning since the novel classes are visually similar to seen classes. In these experiments we compare the performance of the proposed method with three novel class discovery methods, DTC\cite{han2019learning}, RankStat \cite{Han2020Automatically}, and UNO\cite{fini2021unified}. We report our results in Tab.~\ref{tab:finegrained}. Once again our method outperforms all three methods on these fine-grained classification datasets by a significant margin. To be specific, in overall, the proposed method achieves 50-100\% relative improvement compared to the second best method UNO. Together, our previous results combined with these fine-grained results, showcase the efficacy of our proposed method and indicate a wider application for more practical settings.

\begin{table}[ht]
\vspace{-6mm}
\caption{Ablataion study on \textbf{CIFAR-10}, \textbf{CIFAR-100}, and \textbf{Tiny ImageNet} datasets with 50\% classes as seen and 50\% classes as novel. Here, \textbf{UTS} refers to uncertainty-guided temperature scaling, \textbf{MPL} refers to mixed pseudo-labeling, and \textbf{Oracle} refers to having prior knowledge about the number of novel classes.}

\begin{center}
\small
\begin{tabular}{ccc|ccc|ccc|ccc}
\hline


\multicolumn{1}{c}{\multirow{2}{*}{\textbf{UTS}}} & \multicolumn{1}{c}{\multirow{2}{*}{\textbf{MPL}}} & \multicolumn{1}{c|}{\multirow{2}{*}{\textbf{Oracle}}} & \multicolumn{3}{c|}{\textbf{CIFAR-10}} &\multicolumn{3}{c|}{\textbf{CIFAR-100}} &\multicolumn{3}{c}{\textbf{Tiny ImageNet}}\\
\multicolumn{1}{c}{} & \multicolumn{1}{c}{} & \multicolumn{1}{c|}{} & \textbf{Seen} & \textbf{Novel} & \textbf{All} & \textbf{Seen} & \textbf{Novel} & \textbf{All} & \textbf{Seen} & \textbf{Novel} & \textbf{All}

 \\



\hline
\xmark & \xmark & \cmark & $96.0$ & $84.4$ & $90.2$ & $69.2$ & $46.5$ & $57.9$ & 38.1 & 17.5 & 28.1\\
\cmark & \xmark & \cmark & $95.0$ & $86.6$ & $90.8$ & $69.4$ & $46.6$ & $57.9$ & 41.3 & 16.0 & 29.2\\ 
\xmark & \cmark & \cmark & $95.8$ & $87.9$ & $91.9$ & $66.9$ & $48.1$ & $57.5$ & 34.9 & 21.0 & 28.2\\ 
\cmark & \cmark & \xmark & $94.9$ & $89.6$ & $92.2$ & $65.5$ & $44.2$ & $54.8$ & ${40.3}$ & ${19.3}$ & ${30.2}$\\ 
\cmark & \cmark & \cmark & $94.9$ & $89.6$ & $92.2$ & $68.5$ & $52.1$ & $60.3$ & ${39.5}$ & ${20.5}$ & ${30.3}$\\ \hline 


\end{tabular}
\end{center}
\label{tab:ablation}
\vspace{-10mm}
\end{table}
\subsection{Ablation and Analysis}
\label{sec:ablation_analysis}
To investigate the impact of different components, we conduct extensive ablation study on CIFAR-10, CIFAR-100, and Tiny ImageNet datasets. We report the results in Tab.~\ref{tab:ablation}. The first row depicts the performance of our proposed method without uncertainty-guided temperature scaling, and mixed pseudo-labeling. Here, we can see that our proposed method is able to achieve reasonable performance solely based on distribution-aware pseudo-labels. Next, we investigate the impact of removing mixed pseudo-labeling. We observe that the performance on novel classes drops considerably; 3\% on CIFAR-10, 5.5\% on CIFAR-100, and 4.5\% on the Tiny ImageNet dataset. This shows that mixed pseudo-labeling encourages confident learning for novel classes and is a crucial component of our method. After that, we investigate the effect of uncertainty-guided temperature scaling. We observe that the overall performance on all three datasets drops from 0.3\%-2.8\%. We also observe that the performance degradation is more severe on harder datasets (6.9\% relative degradation on Tiny ImageNet compared to 4.6\% on CIFAR-100). Next, we report scores with the estimated number of novel classes (Sec.~\ref{par:novel_estimation}) for completeness (without Oracle in Tab.~\ref{tab:ablation}). We observe that even with the estimated number of novel classes, our method greatly outperforms ORCA and UNO. Our ablation study as a whole demonstrates that every component of our proposed method is crucial and makes a noticeable contribution to the final performance while achieving their designated goal.

\begin{table}
\caption{Estimation of the number of novel classes. The table shows the estimated number of classes vs the actual number of classes in different datasets.}
\begin{center}
\small
\begin{tabular}{l@{\hskip5pt}c@{\hskip5pt}|@{\hskip5pt}c@{\hskip5pt}|@{\hskip5pt}c}
\hline

\textbf{Dataset}  & \textbf{GT} & \textbf{Estimated} & \textbf{Error}\\


\hline
CIFAR-10 & $10$ & $10$ & $0\%$\\
CIFAR-100 & $100$ & $117$ & $17\%$\\
ImageNet-100 & $100$ & $139$ & $39\%$\\
Tiny ImageNet & $200$ & $192$ & $-4\%$\\
\hline 


\end{tabular}
\end{center}
\label{tab:estimation}
\vspace{-6mm}
\end{table}

\begin{wrapfigure}{R}{0.5\textwidth}
\vspace{-10mm}
\begin{center}
  \includegraphics[width=\linewidth]{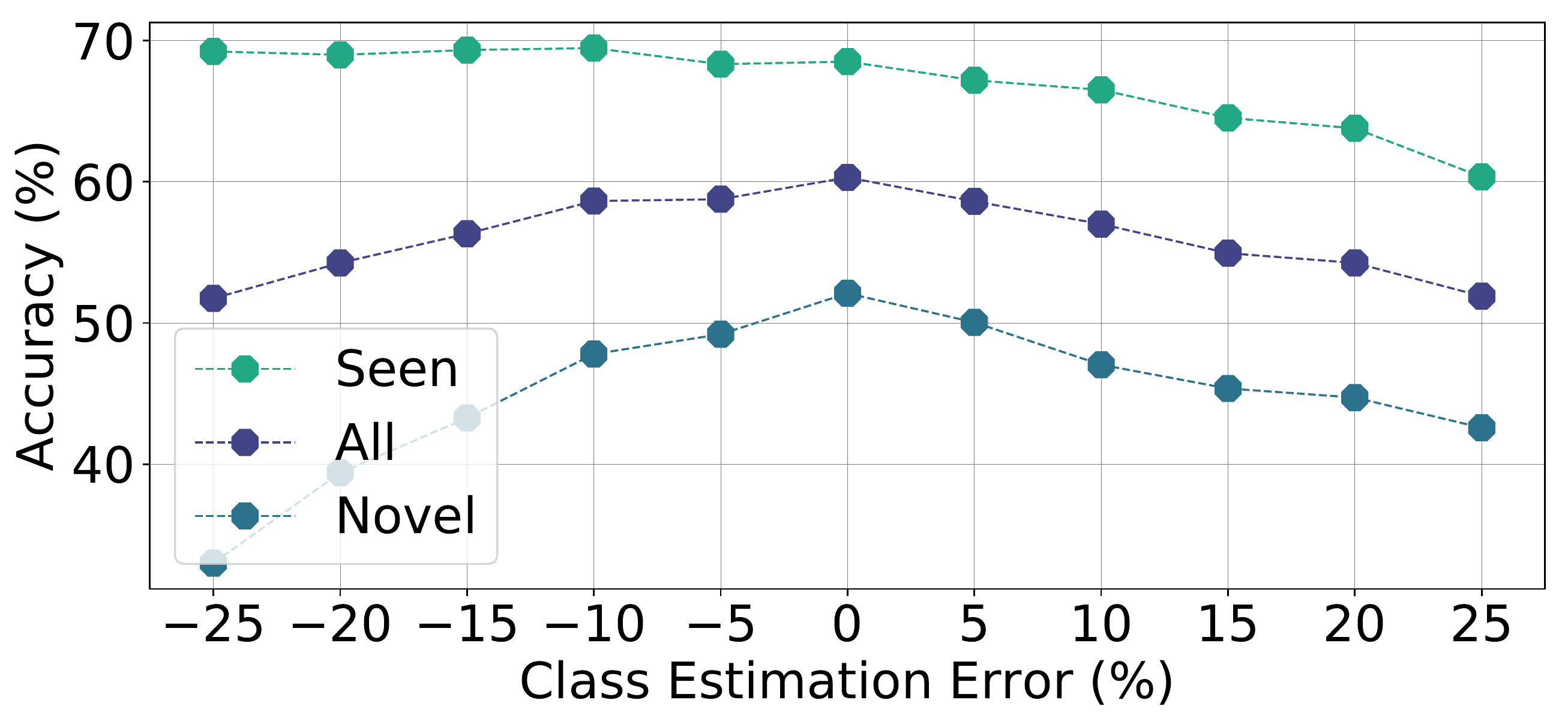}
\end{center}
\vspace{-6mm}
\caption{Accuracy as a function of class estimation error on \textbf{CIFAR-100} dataset.}
\vspace{-4mm}
\label{fig:sensitivity}
\end{wrapfigure}

\vspace{1mm}
\label{par:novel_estimation}
\noindent \textbf{Estimating Number of Novel Classes} A realistic semi-supervised learning system should make minimal assumption about the nature of the problem. For open-world SSL problem, determining the number of novel classes is a crucial step since without explicit determination of the number of classes either a method will have to assume that the number of novel classes is known in advance or set an upper limit for the number of novel classes. A more practical approach is to estimate the number of unkown classes. Therefore, this work proposes a solution to explicitly estimate the number of novel classes. To this end, we leverage self-supervised features from SimCLR \cite{chen2020simple}. 

To estimate the number of novel classes, we perform $k$-means clustering on SimCLR features with different values of $k$. We determine the optimal $k$ by evaluating the performance of generated clusters on the labeled samples. We empirically find that this approach generally underestimates the number of novel classes. This is to be expected since clustering accuracy usually decreases with increasing number of clusters due to assignment of labeled samples to unknown clusters. To mitigate this issue, we perform a sample reassignment technique, where we reallocate the labeled samples assigned to unknown clusters (misclassified samples) to their nearest labeled clusters based on their distance from the cluster centers. Additional details are provided in the supplementary materials.

We report the performance of our estimation method in Tab.~\ref{tab:estimation}. We observe that on all four datasets our proposed estimation method leads to reasonable performance. In addition to this, we conduct a series of experiments on CIFAR-100 dataset to determine the sensitivity of the proposed method to the novel class estimation error. The results are reported in Fig.~\ref{fig:sensitivity} where we observe that our proposed method performs reasonably well over a wide range of estimation error. Please note that even with 25\% overestimation and underestimation errors, our proposed method outperforms ORCA and UNO (Tab.~\ref{tab:vision}). These results reaffirms the practicality of the proposed solution.

\begin{table*}[t]
\caption{Performance on \textbf{CIFAR-100} dataset with different imabalance factors (\textbf{IF}) with 50\% classes as seen and 50\% classes as novel.}
\begin{center}
\small
\begin{tabular}{lccc|ccc}
\hline


 \multicolumn{1}{l}{\multirow{2}{*}{\textbf{Method}}} &
 \multicolumn{3}{c|}{\textbf{IF=10}} & \multicolumn{3}{c}{\textbf{IF=20}}\\  
\multicolumn{1}{c}{} & \textbf{Seen} & \textbf{Novel} & \textbf{All} & \textbf{Seen} & \textbf{Novel} & \textbf{All}\\


\hline
Balanced Class Distribution Prior & $48.4$ & $28.6$ & $38.9$ & $44.4$ & $22.9$ & $33.8$\\
Imbalanced Class Distribution Prior & $50.5$ & $30.8$ & $41.0$ & $48.8$ & $24.6$ & $36.9$\\
Estimated Class Distribution Prior & $50.2$ & $31.3$ & $41.3$ & $44.2$ & $24.0$ & $35.3$\\
\hline 


\end{tabular}
\end{center}
\vspace{-4mm}
\label{tab:imbalance}
\vspace{-3mm}
\end{table*}
\vspace{1mm}
\label{par:data_imbalance}
\noindent \textbf{Data Imbalance} Even though most standard benchmark vision datasets are class balanced, in real-world this is hardly the case. Instead, real-world data often demonstrates long-tailed distribution. Since our proposed method can take any arbitrary distribution into account for generating pseudo-labels, it can naturally take imbalance into account. To demonstrate the effectiveness of our proposed method on imbalanced data, we conduct experiments on CIFAR-100 dataset and report the results in Tab.~\ref{tab:imbalance}. We observe that for both imbalance factors (exponential) of 10 and 20, our proposed method with imbalance class distribution prior improves over the balanced prior baseline by 1.1\% and 3.1\%, respectively. We also conduct another set of experiments where we assume no access to class distribution prior. 
To this end, we propose a simple extension of our method to address imbalance problem. In cases where we do not have access to the prior information about the distribution of classes, to train our model, we start with a class-balanced prior. Next, we iteratively update our prior based on the latest posterior class distribution after every few epochs. The results are reported in the last row of Tab.~\ref{tab:imbalance}. We observe that our simple estimation technique performs reasonably well and outperforms the class-balanced baseline with a noticeable margin. In summary, these experiments validate that our proposed method can effectively take advantage of underlying data distribution and work reasonably well even when we do not have access to the class distribution prior.

\begin{figure}
\captionsetup[subfloat]{labelformat=empty}
\vspace{-6mm}
    \centering
    \subfloat[]{{\includegraphics[width=0.33\textwidth]{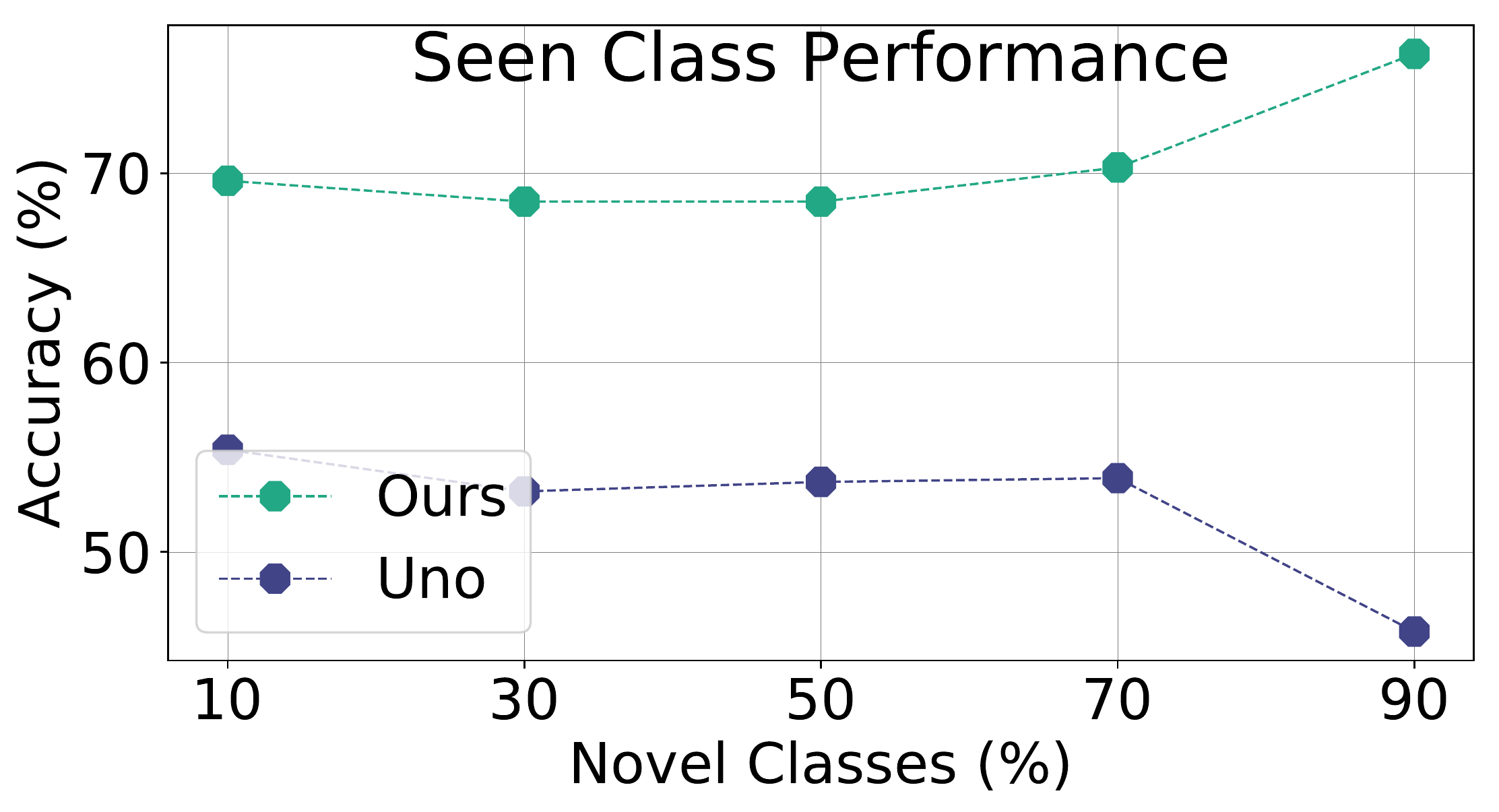}{\label{fig:known}} }}
    \subfloat[]{{\includegraphics[width=0.33\textwidth]{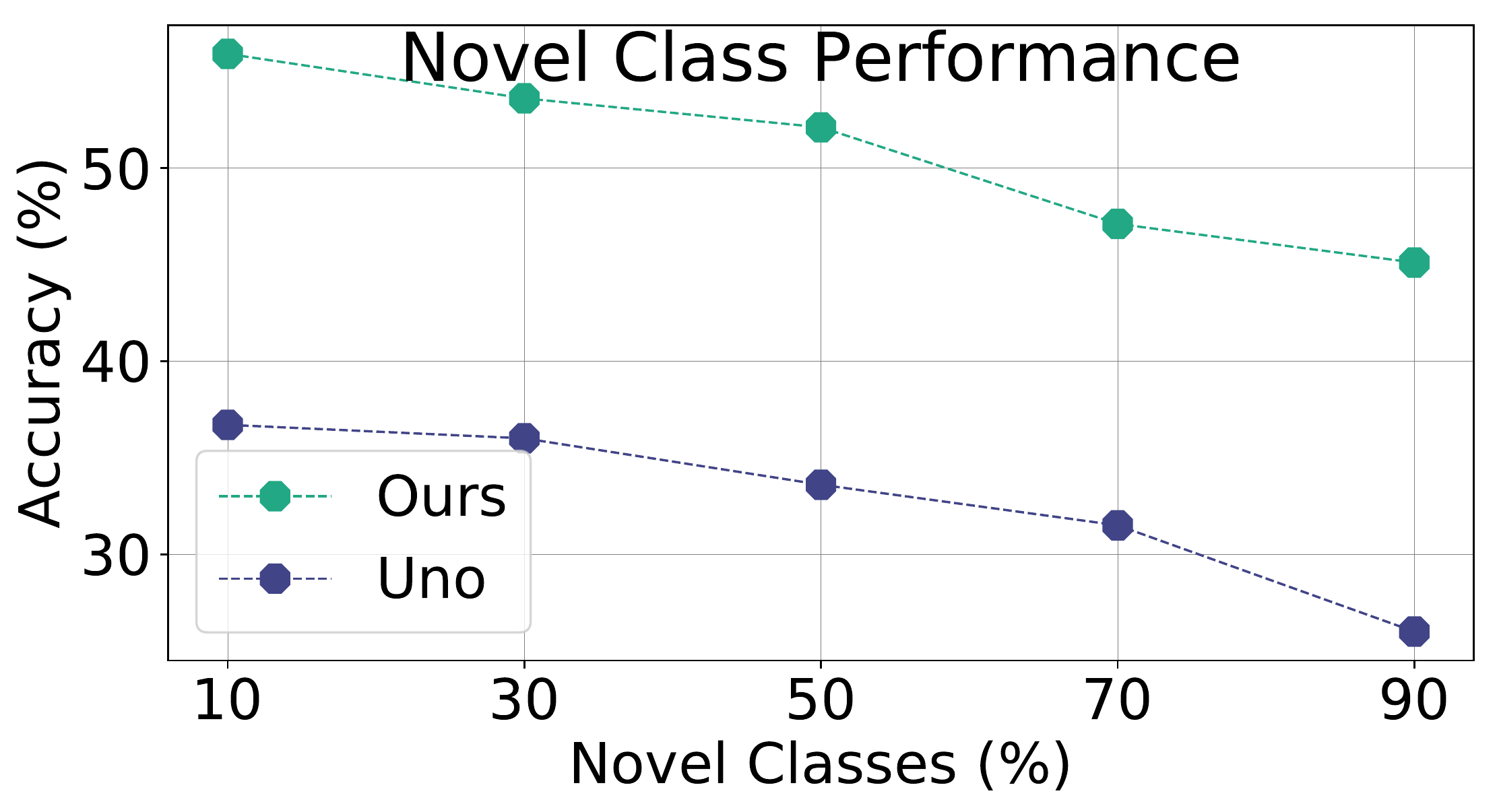}{\label{fig:novel}} }}
    \subfloat[]{{\includegraphics[width=0.33\textwidth]{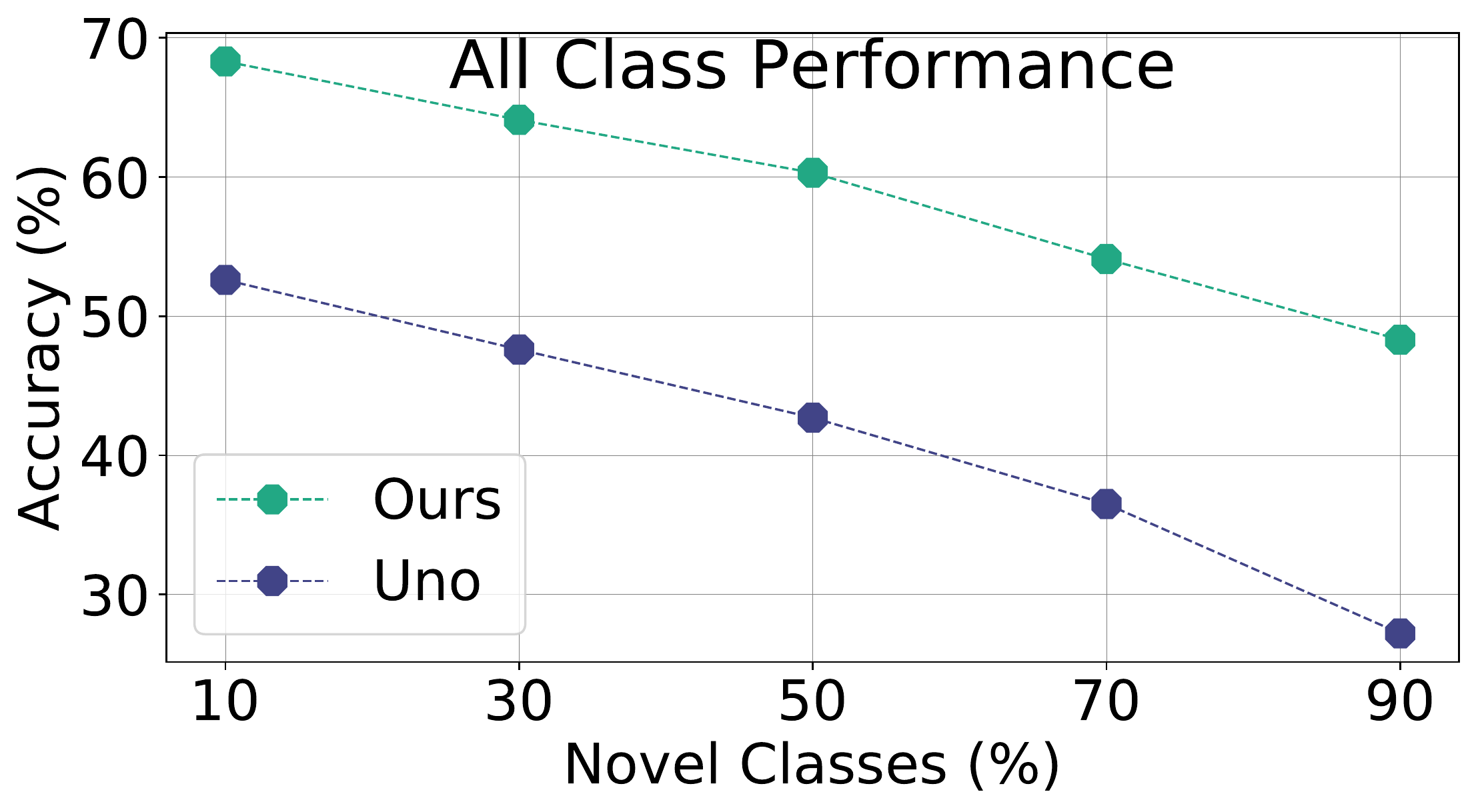}{\label{fig:all}} }}
\vspace{-6mm}
    \caption{Accuracy on seen (left), novel (middle), and all classes (right), as a function of  different percentage of novel classes on the \textbf{CIFAR-100} dataset.}
    \label{fig:differentnovelprct}
    \vspace{-4mm}
\end{figure}

\noindent \textbf{Different Percentage of Novel Classes} 
In all of our experiments, we consider 50\% classes as seen and the remaining 50\% as novel. To further investigate how our method performs under different conditions, we vary percentages of novel classes. We conduct this experiment on CIFAR-100 dataset. The results are presented in Fig.~\ref{fig:differentnovelprct}, where we vary the number of novel classes from 10\% to 90\%. For this analysis, we compare the performance with UNO. The left figure in Fig.~\ref{fig:differentnovelprct} shows that our performance on seen classes remains relatively the same as we increase the percentage of novel classes. Furthermore, we observe that our seen class accuracy increases considerably when the percentage of novel classes is very high (90\%) which is to be expected since this is an easier classification task for seen classes. However, for UNO, we notice a significant performance drop as the number of novel classes increases which shows that UNO is not sufficiently stable for this challenging setup. On novel classes (Fig.~\ref{fig:differentnovelprct}-middle), as we expect, we observe a steady drops in performance as the number of novel classes increase. However, as depicted in this graph, even at a very high novel class ratio, our proposed method can successfully provide a very good performance. Note that, we do not include ORCA in this experiment since their code is not publicly available. However, a similar analysis for ORCA is available in their supplementary materials with 50\% labeled data. We observe that our novel class performance is noticeably higher than ORCA even though we only employ 10\% labeled data. Finally, in Fig.~\ref{fig:differentnovelprct}-right we observe that the overall performance degrades predictably as we increase the percentage of of novel classes.


\begin{wraptable}{R}{0.5\textwidth}
\vspace{-8mm}
\caption{Performance on novel class discovery task on \textbf{CIFAR-100} dataset with 50\% classes as seen and 50\% classes as novel.}
\vspace{-6mm}
\begin{center}
\small
\begin{tabular}{lc}
\hline

\textbf{Method}  & \textbf{Novel}\\

\hline
$k$-means & $28.3$\\
DTC\cite{han2019learning} & $35.9$\\
RankStats\cite{Han2020Automatically} & $39.2$\\
RankStats+\cite{Han2020Automatically} & $44.1$\\
UNO\cite{fini2021unified} & $52.9$\\
Ours & {\cellcolor{yellow!15}}${57.5}$\\\hline 

\end{tabular}
\end{center}
\label{tab:ncd}
\vspace{-8mm}
\end{wraptable}

\vspace{2mm}
\noindent \textbf{Novel Class Discovery}
In this work, we propose a general solution for open-world SSL problem which can be easily modified for the novel class discovery problem, where the principal assumption is that the unlabeled data contains only novel class samples. In this set of experiments we apply our proposed method on the novel class discovery task by generating pseudo-labels only for novel classes. We do not make any other modification to the original method for this task. The findings from these experiments are reported in Tab.~\ref{tab:ncd}. We conduct experiments on CIFAR-100-50, i.e., 50 classes are set as novel. For comparison, we use the results reported in UNO \cite{fini2021unified}. To the best of our knowledge, UNO reports the best scores for this particular experimental setup. Tab.~\ref{tab:ncd} demonstrates that the porposed method outperforms $k$-means, DTC \cite{han2019learning}, RankStats \cite{Han2020Automatically}, and RankStats+ by a significant margin. Importantly, our method also outperforms the current state-of-the-art method for novel class discovery, UNO, by 4.6\%. Interestingly, this experiment demonstrates that \emph{our proposed method is a versatile solution which can be readily applied to novel class discovery problem.}

\section{Conclusion}
In this work, we propose a practical method for open-world SSL problem. Our proposed method generates pseudo-labels according to class distribution prior to solve open-world SSL problem in realistic settings with arbitrary class distributions. We extend our method to handle practical scenarios where neither the number of unkown classes nor the class distribution prior is available. Furthermore, we introduce uncertainty-guided temperature scaling to improve the reliability of pseudo-label learning. Our extensive experiments on seven diverse datasets demonstrate the effectiveness of our approach, where it significantly improves the state-of-the-art. Finally, we show that our method can be readily applied to novel class discovery problem to outperform the existing solutions.

\clearpage
%
%


\appendix
\section*{Appendix}
This appendix includes the following sections. First, our training algorithm is introduced in section \ref{sec:alogrithm}. Then, we provide implementation details in section \ref{sec:implementation}. The details of the datasets that we use in our experiments are available in section \ref{sec:dataset}. Next, in sections \ref{sec:morelabeled} and \ref{sec:limited}, we provide additional results with varying the amount of labeled data. After that, we provide additional details about estimating the number of novel classes in section \ref{sec:novel_estimate}. We discuss the effect of varying the number of novel classes for the novel class discovery (NCD) task in section \ref{sec:novel}. Next, we analyse the effect of changing temperature in section \ref{sec:temp}. Finally, in section \ref{sec:conf_mat}, we demonstrate that our proposed method is able to recognise novel classes without confusing them with seen classes. 

\section{Training Algorithm}
\label{sec:alogrithm}

We provide our training algorithm in Alg.~\ref{alg: alogirthm1}. For training, we require access to a labeled dataset, $\mathbb{D}_L$, and a set of unlabeled data, $\mathbb{D}_U$. We also require the prior class distribution, $\rho = \langle N_{U}^{C_1}/{N_U,...,N_{U}^{C_{|\mathbb{C}_L|+|\mathbb{C}_N|}}}/N_U \rangle$, iterations per epoch, $E$, the maximum iterations, $K$, and the temperature, $T$. First, we initialize the neural network, $f_w$, and assign $T$ to the sample uncertainties, $\mathbb{U}_L$ and $\mathbb{U}_U$. Next, we sample a batch of labeled and unlabeled data, as well as their corresponding sample uncertainties. After that, we obtain the class-distribution-aware pseudo-labels using Sinkhorn-Knopp algorithm\cite{sinkhorn1967concerning} while minimizing the optimization problem in Eq.~\ref{eqn:minimize}. 

\begin{align}
\label{eqn:minimize}
    \min_{\mathbf{A}\in\mathcal{A}_\rho}-Tr((\mathbf{A}\mathbf{P_\pi})^T\log(\mathbf{\hat{Y}}_U/N_U)),
\end{align}
where $\mathcal{A}_\rho$ is the transportation polytope defined in Eq.~3 of the main text which satisfies the prior class distribution $\rho$, $P_{\pi}$ is the permutation matrix that reorders the columns of the assignment matrix, $\mathbf{A}$, according to the order of marginals of the output probabilities, $\mathbf{\hat{Y}}$.

We also perform cross pseudo-labeling to encourage perturbation invariant feature learning. Next, we generate the hard pseudo-labels for confident novel classes from the generated pseudo-labels. As a result we will have a mixture of soft and hard pseudo-labels. Then, we concatenate the input images from the labeled and unlabeled batch. We also concatenate the ground-truth labels and generated coherent pseudo-labels, and the sample uncertainty scores. In the next step, we  perform Mixup augmentation\cite{zhang2018mixup} on the concatenated inputs, labels, and uncertainty scores. Once we obtain the mixed inputs, labels, and uncertainty values, we update the network parameters using cross entropy loss. We update the uncertainty values for the unlabeled samples at the end of each epoch ($E$ iterations). Finally, the algorithm ends after completing $K$ iterations and returns the trained neural network, $f_w$.       

\begin{algorithm*}[h]
\caption{Training algorithm}
\label{alg: alogirthm1}
\textbf{Input:} Labeled data, $\mathbb{D}_L$, a set of unlabeled data, prior class distribution $\rho$, $\mathbb{D}_U$, iterations per epoch $E$, maximum iterations $K$, temperature $T$.  \\
\textbf{Output:} Trained neural network, $f_{w}$.
\begin{algorithmic}[1]
\State Initialize neural network, $f_{w}$.
\State $\mathbb{U}_L \gets T$, $\mathbb{U}_U \gets T$ \Comment{$\mathbb{U}_L$, and $\mathbb{U}_U$ are sample uncertainties.}
\For{$k$ = 1...{$K$}} 
    \State ($\mathbf{X}_l, \mathbf{Y}_l, \mathbf{u}_l)  \gets \mathrm{SampleBatch}(\mathbb{D}_L, \mathbb{U}_L)$
    \State ($\mathbf{X}_u, \mathbf{X'}_u, \mathbf{u}_u) \gets\mathrm{SampleBatch}(\mathbb{D}_U, \mathbb{U}_U)$
  \State $\mathbf{\tilde{Y}}_u, \mathbf{\tilde{Y}'}_u \gets \mathrm{Sinkhorn}(f_w(\mathbf{X'}_u), \rho), \mathrm{Sinkhorn}(f_w(\mathbf{X}_u), \rho)$  \Comment{Eq.~\ref{eqn:minimize} and cross PL.}
  \State $\mathbf{\Bar{Y}}_u, \mathbf{\Bar{Y}'}_u  \gets \mathrm{MixedPL}(\mathbf{\tilde{Y}}_u), \mathrm{MixedPL}(\mathbf{\tilde{Y}'}_u)$
  \State $\mathbf{{X}}, \mathbf{{Y}}, \mathbf{{u}} \gets \mathrm{Concat}(\mathbf{X}_l, \mathbf{X}_u, \mathbf{X'}_u), \mathrm{Concat}(\mathbf{Y}_l, \mathbf{\Bar{Y}}_u, \mathbf{\Bar{Y'}}_u), \mathrm{Concat}(\mathbf{u}_l, \mathbf{{u}}_u, \mathbf{{u}}_u)$
  \State $\mathbf{{X}}_m, \mathbf{{Y}}_m, \mathbf{{u}}_m \gets \mathrm{Mixup}(\mathbf{{X}}, \mathbf{{Y}}, \mathbf{{u}})$
  \State ${w}^{(k+1)} \gets w^{(k)} - \alpha\nabla_{w}\mathcal{L}_{ce}(w^{(k)}, \mathbf{{X}}_m, \mathbf{{Y}}_m, \mathbf{{u}}_m)$
  \If{$k\%E=0$} 
  \State $\mathbb{U}_U \gets \mathrm{Uncertainty}(\mathbb{D}_U)$ \Comment{Eq.~6}
  \EndIf
 \EndFor

\State \textbf{return} $f_{w}$

\end{algorithmic}
\end{algorithm*}


\section{Implementation Details}
\label{sec:implementation}
To effectively process the lower resolution images from CIFAR-10 and CIFAR-100 datasets, similar to previous works \cite{cao2022openworld,Han2020Automatically,fini2021unified}, we modify the first convolutional layer and set the kernel size to 3$\times$3 and apply a stride of 1. In addition, we remove the first max-pooling layer. We make a similar change for experiments in Tiny ImageNet dataset. However, we do not remove the first max-pooling layer since the images are of higher resolution. For ImageNet-100 and the fine-grained dataset experiments we do not make any changes to the network. Besides, for comparison, we use the same network for all the methods.       

For data augmentation, we primarily use SimCLR\cite{chen2020simple} augmentations, which include: random resized crop, horizontal flip, color jittering, random grayscale, and Gaussian blur. For CIFAR-10 and CIFAR-100 we use solarize and equalize transformations instead of random grayscale and Gaussian blur. We also use Mixup augmentation in our training. For Mixup \cite{zhang2018mixup} augmentation, $\gamma$ is set to $0.75$. For all of our experiments, we use a threshold of $0.5$ for generating hard pseudo-labels for the confident novel class samples.     

For all of our experiments, except CIFAR-10 and CIFAR-100 experiments with 50\% labeled data (Sec.~\ref{sec:morelabeled}), we apply a temperature value of 0.1. In our experiments, we observe that higher number of labeled examples per class (CIFAR-10 and CIFAR-100) creates a relatively stronger bias towards known classes when the temperature value is low. To address this issue, we use a temperature of 0.2 for the 50\% labeled data experiments on CIFAR-10 and CIFAR-100 datasts. For performing uncertainty-guided temperature scaling we normalize the uncertainty values of the entire dataset to make the maximum uncertainty value 1. Finally, we clip the uncertainty values between 0.1 and 1.0 so that very low uncertainty values do not lead to overconfident predictions. 
\begin{table*}[h]
\caption{Details of the datasets used in our experiments.}
\label{tab:dataset}
\begin{center}\setlength{\tabcolsep}{4pt}
\small
\begin{tabular}{lccc}
\hline


\textbf{Dataset} & \textbf{No Class} & \textbf{Train Samples} & \textbf{Test Samples}
 \\


\hline
CIFAR-10~\cite{cifar10} & $10$ & $50,000$ & $10,000$ \\
CIFAR-100~\cite{cifar100} & $100$ & $50,000$ & $10,000$ \\
ImageNet-100~\cite{russakovsky2015imagenet} & $100$ & $128,545$ & $5,000$ \\
Tiny ImageNet~\cite{le2015tiny} & $200$ & $100,000$ & $10,000$\\
Oxford-IIIT Pet~\cite{parkhi12a} & $37$ & $3,680$ & $3,669$\\
FGVC-Aircraft~\cite{maji13fine-grained} & $100$ & $6,667$ & $3,333$ \\
Stanford-Cars~\cite{KrauseStarkDengFei-Fei_3DRR2013} & $196$ & $8,144$ & $8,041$\\ \hline 


\end{tabular}
\end{center}
\end{table*}

\section{Datasets}
\label{sec:dataset}
We provide the details of the datasets used in our experiments in Tab.~\ref{tab:dataset}, which shows the number of classes in each dataset alongside the number of train and test samples. For FGVC-Aricraft\cite{maji13fine-grained} dataset, we train our model on the joint set of training and validation samples. Besides, since Oxford-IIIT Pet dataset contains odd number of classes, in 50\% novel class experiment, we treat the first 19 classes of this dataset as seen and the remaining 18 classes as novel.

The input resolution of CIFAR-10 and CIFAR-100 images is 32$\times$32; Tiny ImageNet images are slightly larger, i.e., 64$\times$64. For the fine-grained datasets the images vary in size and aspect ratio. Therefore, for computational efficiency, we pre-process the images for fine-grained datasets and resize them to 256$\times$256 resolution; this pre-processing operation is performed for both train and test images in all of our experiments.

\begin{table}[h]
\caption{Accuracy on \textbf{CIFAR-10} and \textbf{CIFAR-100} datasets with 50\% classes as seen and 50\% classes as novel.}
\label{tab:cifar10_cifar100}
\begin{center}\setlength{\tabcolsep}{4pt}
\small
\begin{tabular}{lccc|ccc}
\hline


 \multicolumn{1}{l}{\multirow{2}{*}{\textbf{Method}}} & \multicolumn{3}{c|}{\textbf{CIFAR-10}} & \multicolumn{3}{c}{\textbf{CIFAR-100}} \\  
\multicolumn{1}{c}{} & \textbf{Seen} & \textbf{Novel} & \textbf{All}  & \textbf{Seen} & \textbf{Novel} & \textbf{All}\\


\hline
FixMatch\cite{sohn2020fixmatch} & $71.5$ & $50.4$ & $49.5$ & $39.6$ & $23.5$ & $20.3$\\
DS$^{3}$L\cite{guo2020safe} & $77.6$ & $45.3$ & $40.2$ & $55.1$ & $23.7$ & $24.0$\\
CGDL\cite{sun2020conditional} & $72.3$ & $44.6$ & $39.7$ & $49.3$ & $22.5$ & $23.5$\\
DTC~\cite{han2019learning} & $53.9$ & $39.5$ & $38.3$ & $31.3$ & $22.9$ & $18.3$\\
RankStats\cite{Han2020Automatically} & $86.6$ & $81.0$ & $82.9$ & $36.4$ & $28.4$ & $23.1$\\
SimCLR\cite{chen2020simple} & $58.3$ & $63.4$ & $51.7$ & $28.6$ & $21.1$ & $22.3$\\
UNO\cite{fini2021unified} & $91.6$ & $69.3$ & $80.5$ & $68.3$ & $36.5$ & $51.5$\\
ORCA\cite{cao2022openworld} & $88.2$ & $90.4$ & $89.7$ & $66.9$ & $43.0$ & $48.1$\\
Ours & {\cellcolor{yellow!15}}${96.8}$ & {\cellcolor{yellow!15}}${92.8}$ & {\cellcolor{yellow!15}}${94.8}$ & {\cellcolor{yellow!15}}${80.2}$ & {\cellcolor{yellow!15}}${49.3}$ & {\cellcolor{yellow!15}}${64.7}$\\\hline 


\end{tabular}
\end{center}
\vspace{-2mm}
\end{table}

\begin{table}[h]
\caption{Accuracy on \textbf{Tiny ImageNet}  dataset with  50\% labeled data. We consider 50\% classes as seen and 50\% classes as novel.}
\label{tab:cifar100}
\begin{center}\setlength{\tabcolsep}{4pt}
\small
\begin{tabular}{lccc}
\hline

\textbf{Method}  & \textbf{Seen} & \textbf{Novel} & \textbf{All}\\


\hline
DTC~\cite{han2019learning} & $28.8$ & $16.3$ & $19.9$\\
RankStats~\cite{Han2020Automatically} &$5.7$ & $5.4$ & $3.4$\\
UNO~\cite{fini2021unified} &  $46.5$ & $15.7$ & $30.3$\\
Ours & {\cellcolor{yellow!15}}${59.1}$ & {\cellcolor{yellow!15}}${24.2}$ & {\cellcolor{yellow!15}}${41.7}$\\\hline 


\end{tabular}
\end{center}
\vspace{-4mm}

\end{table}

\section{Experiments with More Labeled Data}
\label{sec:morelabeled}
In this work we propose a solution for the realistic open-world SSL problem. Therefore, in the main text, we include experiments with only limited number of labeled examples (10\%). In this section, we provide additional results with more labeled data to provide additional comparison with other methods. To this end, we conduct experiments with 50\% labeled data on CIFAR-10, CIFAR-100, and Tiny ImageNet. Tab.~\ref{tab:cifar10_cifar100} reports the results on CIFAR-10 and CIFAR-100 datasets. We observe that similar to experiments with 10\% labeled data, our proposed method outperforms all the other techniques. To be specific, in parallel to outperforming ORCA\cite{cao2022openworld}, our proposed algorithm also outperforms popular self-supervised learning method, SimCLR\cite{chen2020simple}, and a recently proposed open-set recognition method, CGDL\cite{sun2020conditional}. On CIFAR-10 dataset our porposed method outperforms ORCA\cite{cao2022openworld} by 5.1\% and on CIFAR-100 dataset it outperforms the second best method UNO\cite{fini2021unified} by 13.2\%.

We conduct similar experiments on Tiny ImageNet dataset. Similar to our experiments in the main text, we compare our performance with DTC\cite{han2019learning}, RankStats\cite{Han2020Automatically}, and UNO\cite{fini2021unified}. We observe that our proposed method outperforms the second best method UNO by 11.4\%. The experiments on these three datasets demonstrate that our proposed method can perform reasonably well even with more labeled data.

\begin{table}[h]
\caption{Accuracy on \textbf{CIFAR-10} dataset with 50\% classes as seen and 50\% classes as novel.}
\label{tab:cifar10}
\begin{center}\setlength{\tabcolsep}{4pt}
\small
\begin{tabular}{lccc|ccc}
\hline


 \multicolumn{1}{c}{\multirow{2}{*}{\textbf{Method}}} & \multicolumn{3}{c|}{\textbf{1\% Data}} & \multicolumn{3}{c}{\textbf{5\% Data}} \\  
\multicolumn{1}{c}{} & \textbf{Seen} & \textbf{Novel} & \textbf{All}  & \textbf{Seen} & \textbf{Novel} & \textbf{All}\\


\hline
UNO\cite{fini2021unified} & $48.4$ & $67.1$ & $52.6$ & $74.6$ & $68.4$ & $71.8$\\
Ours & {\cellcolor{yellow!15}}${92.0}$ & {\cellcolor{yellow!15}}${91.1}$ & {\cellcolor{yellow!15}}${91.6}$ & {\cellcolor{yellow!15}}${90.9}$ & {\cellcolor{yellow!15}}${91.4}$ & {\cellcolor{yellow!15}}${91.2}$\\\hline 


\end{tabular}
\end{center}
\vspace{-3mm}
\end{table}

\section{Reducing the Number of Labeled Data}
\label{sec:limited}
In this section, we discuss additional experiments with lower number of labeled data. We report results on CIFAR-10 dataset with only 1\% and 5\% labeled data in Tab.~\ref{tab:cifar10}. Since the source code for ORCA\cite{cao2022openworld} is not publicly available, we restrict our comparison to UNO\cite{fini2021unified}, which is the previous best method on this dataset. We observe that our proposed method significantly outperforms UNO in both of these experimental setups. We also notice that the performance of UNO on seen classes significantly degrades when only 1\% labeled data is available, which is not the case for our porposed algorithm. Furthermore, even though we do not directly compare our results with ORCA and other baseline methods for these challenging experiments, \emph{we notice that our proposed method with 1\% and 5\% labeled data, is able to outperform ORCA with higher number of labeled data, both 10\% and 50\% labeled data, on seen/novel/all class performances.}   

\begin{table}[h]
\vspace{-2mm}
\caption{Accuracy on \textbf{CIFAR-100}  dataset with  5\% labeled data. We consider 50\% classes as seen and 50\% classes as novel.}
\label{tab:cifar100}
\begin{center}\setlength{\tabcolsep}{4pt}
\small
\begin{tabular}{lccc}
\hline

\textbf{Method}  & \textbf{Seen} & \textbf{Novel} & \textbf{All}\\


\hline
UNO\cite{fini2021unified} & $44.0$ & $31.7$ & $36.5$\\
Ours & {\cellcolor{yellow!15}}${60.1}$ & {\cellcolor{yellow!15}}${47.4}$ & {\cellcolor{yellow!15}}${54.4}$\\\hline 


\end{tabular}
\end{center}

\vspace{-3mm}
\end{table}

Next, we conduct experiments on CIFAR-100 dataset with only 5\% labeled data. The results are depicted in Tab.~\ref{tab:cifar100}. For this experiment we also compare our results with UNO\cite{fini2021unified}, which is the previous best method on this dataset. We notice that similar to the results on CIFAR-10, our proposed method outperforms UNO by a large margin and achieves 15.7\% improvement over UNO on novel classes. Besides, the performance on novel classes and all classes is better than ORCA even when ORCA uses 50\% of labeled data. The results on these two datasets demonstrate that our proposed algorithm is much more label efficient than ORCA, and can achieve strong performance even when only a handful of labeled examples (250 labeled examples in CIFAR-10 1\% labeled data experiment) are available.

\begin{table*}[h]
\caption{Accuracy on \textbf{Oxford-IIIT Pet}, \textbf{FGVC-Aircraft}, and \textbf{Stanford-Cars}  datasets with 25\% labeled data. We consider 50\% classes as seen and 50\% classes as novel.}
\label{tab:finegrained}
\begin{center}\setlength{\tabcolsep}{4pt}
\small
\begin{tabular}{lccc|ccc|ccc}
\hline


 \multicolumn{1}{c}{\multirow{2}{*}{\textbf{Method}}} & \multicolumn{3}{c|}{\textbf{Oxford-IIIT Pet}} & \multicolumn{3}{c|}{\textbf{FGVC-Aircraft}} &\multicolumn{3}{c}{\textbf{Stanford-Cars}} \\  
\multicolumn{1}{c}{} & \textbf{Seen} & \textbf{Novel} & \textbf{All}  & \textbf{Seen} & \textbf{Novel} & \textbf{All}  & \textbf{Seen} & \textbf{Novel} & \textbf{All}\\


\hline

UNO\cite{fini2021unified} & $35.6$ & $19.1$ & $25.8$ & $28.2$ & $20.7$ & $21.9$ & $26.5$ & $10.3$ & $17.2$\\
Ours & {\cellcolor{yellow!15}}${59.1}$ & {\cellcolor{yellow!15}}${31.4}$ & {\cellcolor{yellow!15}}${45.6}$ & {\cellcolor{yellow!15}}$52.4$ & {\cellcolor{yellow!15}}${36.5}$ & {\cellcolor{yellow!15}}${45.3}$ & {\cellcolor{yellow!15}}${67.4}$ & {\cellcolor{yellow!15}}${32.5}$ & {\cellcolor{yellow!15}}${50.0}$\\\hline 


\end{tabular}
\end{center}

\end{table*}

Finally, we conduct experiments on the fine-grained datasets with only 25\% labeled data. We choose 25\% labeled data for these experiments since even with such a large portion of labeled data, the number of labeled samples is not greater than $\sim$1000 samples. This experimental setup is challenging, since handling such large resolution images with a large number of classes is always difficult for neural networks. We present the results in Tab.~\ref{tab:finegrained}. We compare our results with UNO\cite{fini2021unified} since the source code for ORCA is not publicly available. The results in Tab.~\ref{tab:finegrained} demonstrates that similar to 50\% labeled data experiment, our proposed method outperforms UNO significantly on all of these datasets. These results further validate the effectiveness of our method by demonstrating that it can work on challenging fine-grained classification tasks with a large number of classes while using only a handful of labeled examples.

\begin{table}
\caption{Estimation of the number of novel classes. The table shows the estimated number of classes on different datasets with and without sample reassignment technique.}
\begin{center}\setlength{\tabcolsep}{4pt}
\small
\begin{tabular}{lccc}
\hline

\multicolumn{1}{l}{\multirow{2}{*}{\textbf{Dataset}}}  & \multicolumn{1}{c}{\multirow{2}{*}{\textbf{GT}}} & \multicolumn{2}{c}{\textbf{Estimated}}\\
\multicolumn{1}{c}{} & \multicolumn{1}{c}{} & \multicolumn{1}{c}{\textbf{w/o reassignment}} & \multicolumn{1}{c}{\textbf{w reassignment}}\\


\hline
CIFAR-10 & $10$ & $10$ & $10$\\
CIFAR-100 & $100$ & $87$ & $117$\\
ImageNet-100 & $100$ & $84$ & $139$\\
Tiny ImageNet & $200$ & $132$ & $192$\\
\hline 


\end{tabular}
\end{center}
\label{tab:estimation}
\vspace{-6mm}
\end{table}

\section{Estimating Number of Novel Classes}
\label{sec:novel_estimate}
For estimating the number of novel classes we vary the value of $k$ from the number of labeled classes to 400. We can potentially use a higher upper limit but our experiment demonstrate that using a higher number does not change the number of estimated classes. For each value of $k$, we average the results from 3 independent runs of the $k$-means clustering algorithm to obtain more stable performance. For sample reassignment, first we perform Hungarian matching \cite{kuhn1955hungarian} for the labeled samples. Such matching provides us the dominant clusters for the labeled samples. After that, we select the misclassified labeled examples and reassign them to these dominant labeled clusters using their nearest neighbor cluster. To obtain the final estimate of the number of classes we average the top 10 values and use that as our estimate to make the prediction more stable.

Table \ref{tab:estimation} reports the performance of our number of class estimation procedure with and without misclassified labeled example reassignment. Overall, we are able to estimate the number of classes reasonably well on the four common benchmark datasets that we investigated in our study. We observe that generally without the reassignment step, the estimated number of classes tends to be lower than the actual number of classes. On the more challenging Tiny ImageNet dataset, the reassignment step seems crucial for obtaining more accurate estimates.

\begin{figure}[h]
\begin{center}
\includegraphics[width=0.6\textwidth]{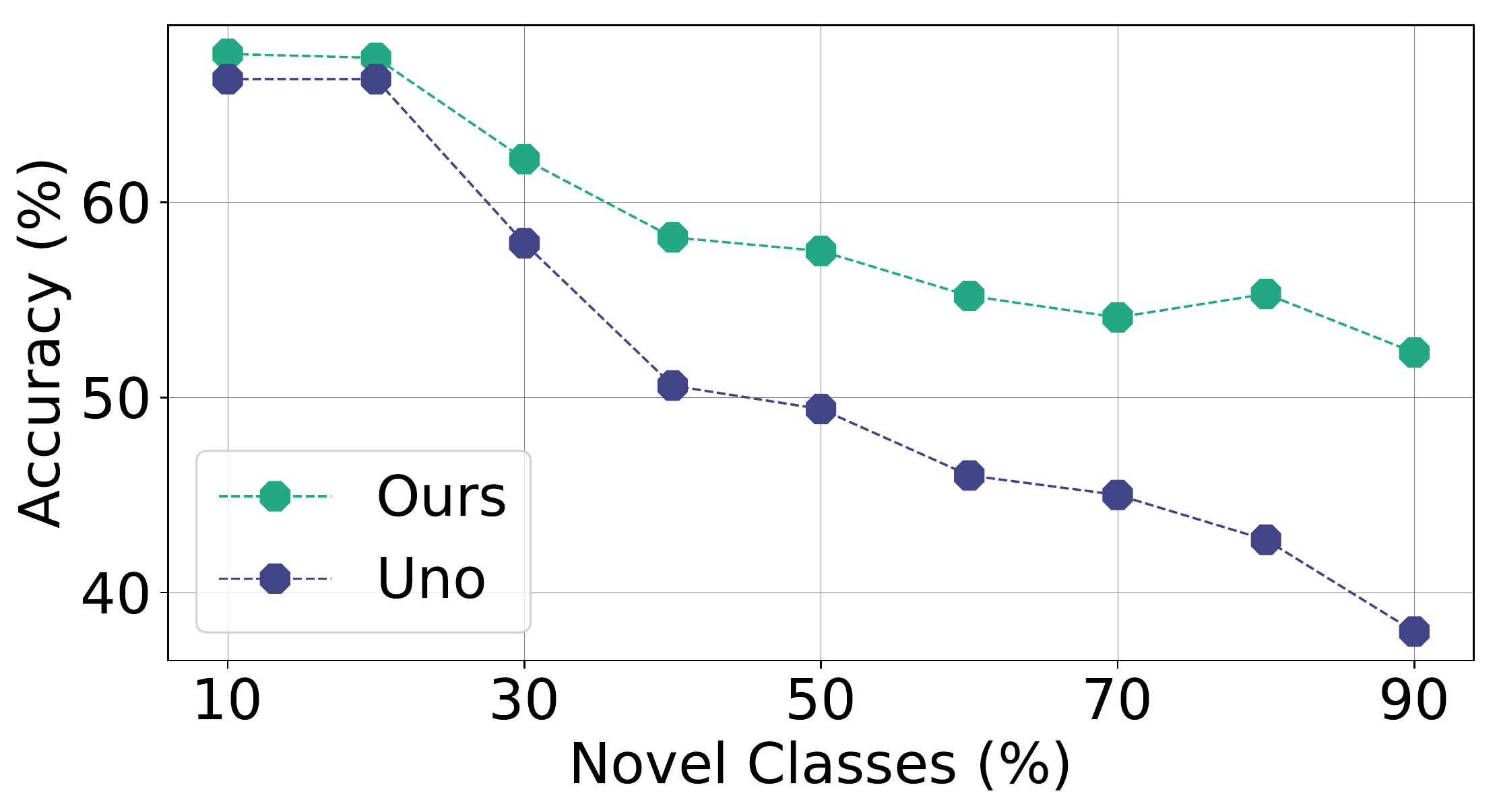}
\caption{Performance on \textbf{CIFAR-100} dataset for novel class discovery (NCD) task with different numbers of novel classes.}
\label{fig:ncd}
\vspace{-12mm}
\end{center}
\end{figure}

\section{Varying the Number of Novel Classes for NCD Task}
\label{sec:novel}
In the main text, we provide results on novel class discovery (NCD) task on CIFAR-100 dataset. In this section, we provide additional results by varying the percentage of novel classes. The results are provided in Fig.~\ref{fig:ncd}. We compare our method only with UNO since it outperforms the previous works with a significant margin. For this comparison we use the official code available for UNO. In this comparison, we report `task-agnostic' accuracy, which is a more realistic evaluation. In `task-agnostic' setting, we assume that we do not have any knowledge about the sample belonging to seen classes or novel classes. We observe that UNO achieves similar performance to our method when the number of novel classes is lower. However, as the percentage of novel classes increase, performance of UNO falls behind significantly. Moreover, we observe a predictable drop in performance for both methods as we increase the percentage of novel classes. In summary, these results demonstrate that our method can work well in more challenging scenarios where the number of novel classes are significantly higher than the seen classes.


\begin{figure*}
\captionsetup[subfloat]{labelformat=empty}
    \centering
    \subfloat[]{{\includegraphics[width=0.32\textwidth]{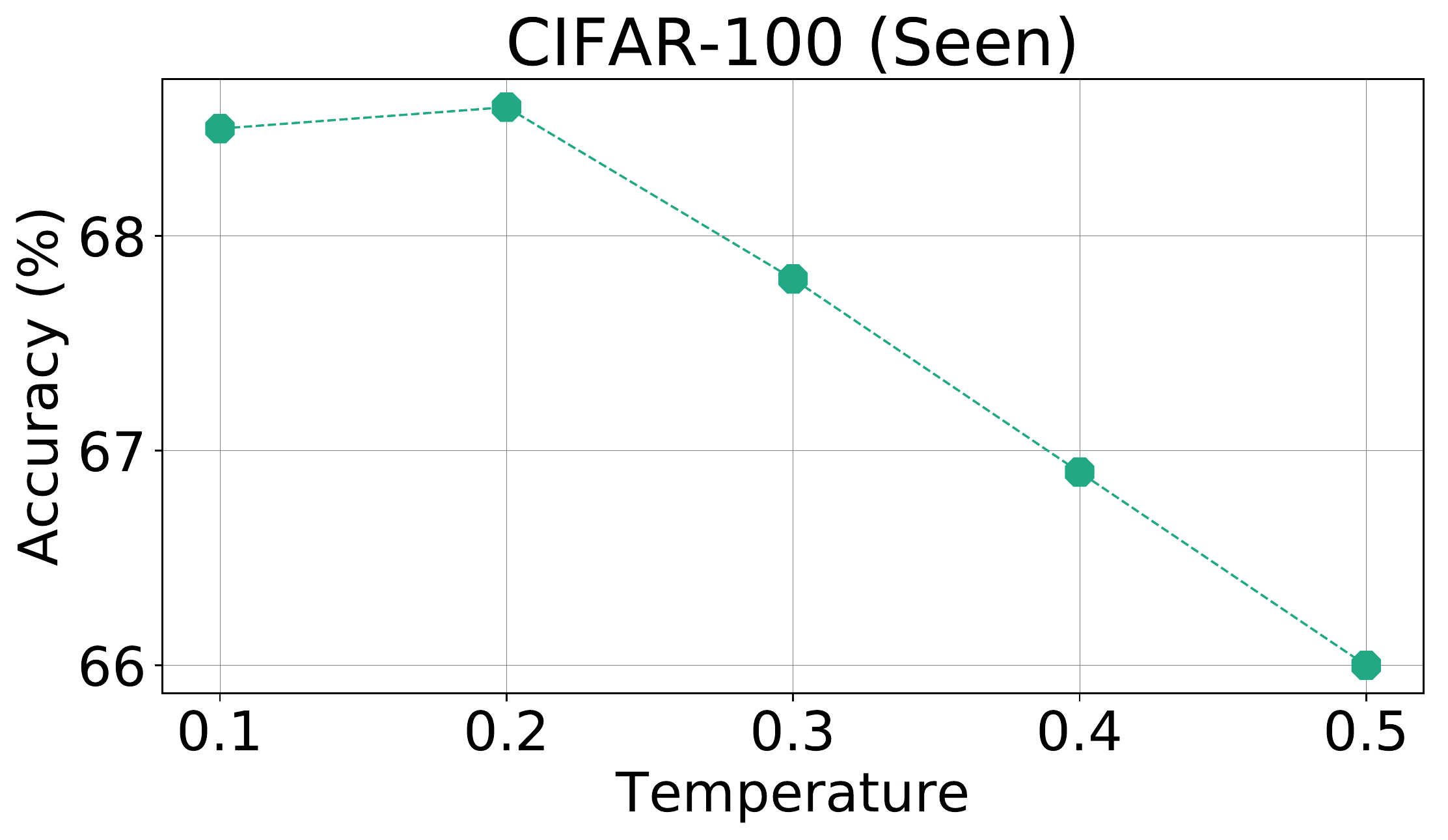}{\label{fig:temp_seen}} }}
    \subfloat[]{{\includegraphics[width=0.32\textwidth]{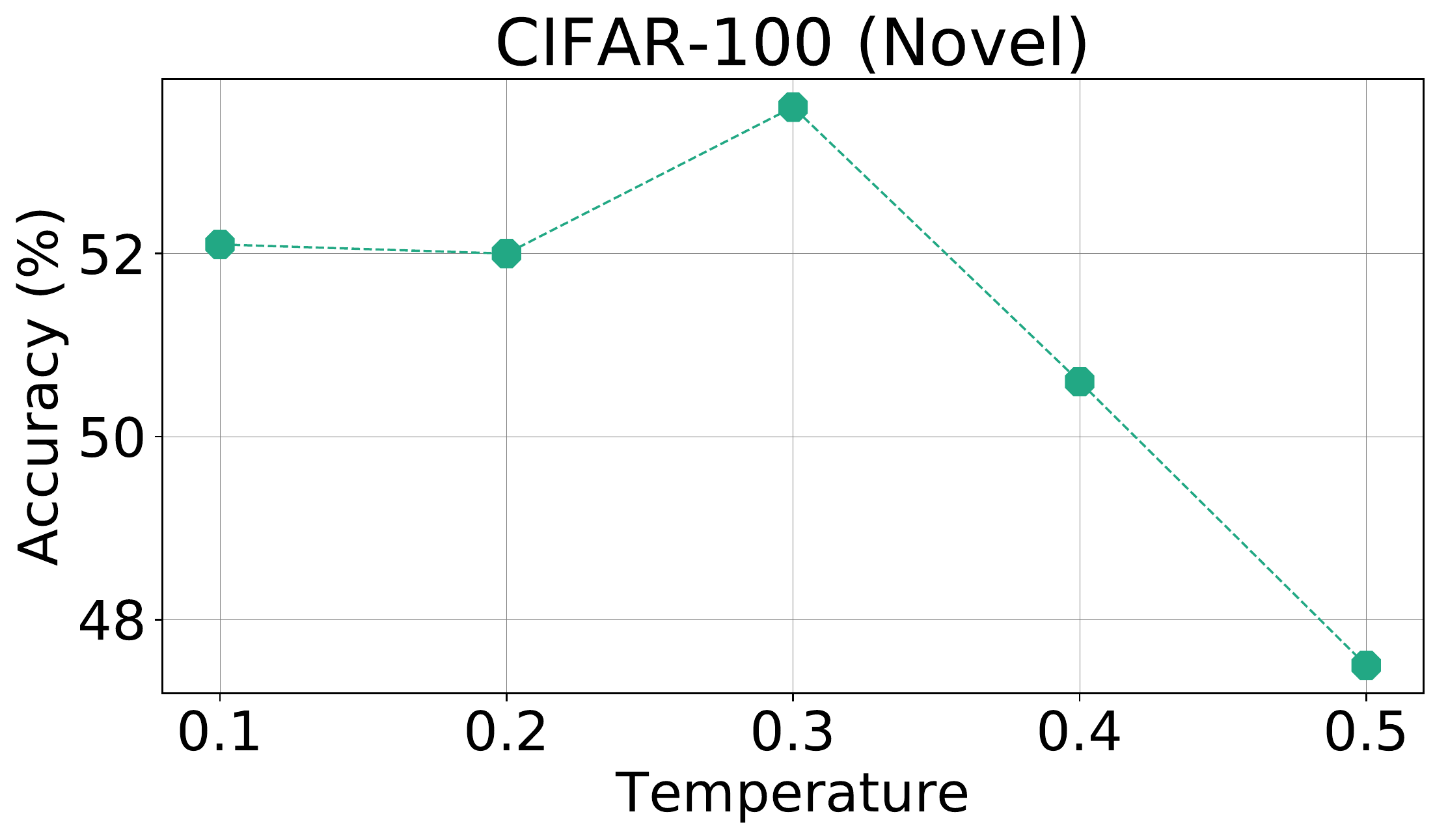}{\label{fig:temp_novel}} }}
    \subfloat[]{{\includegraphics[width=0.32\textwidth]{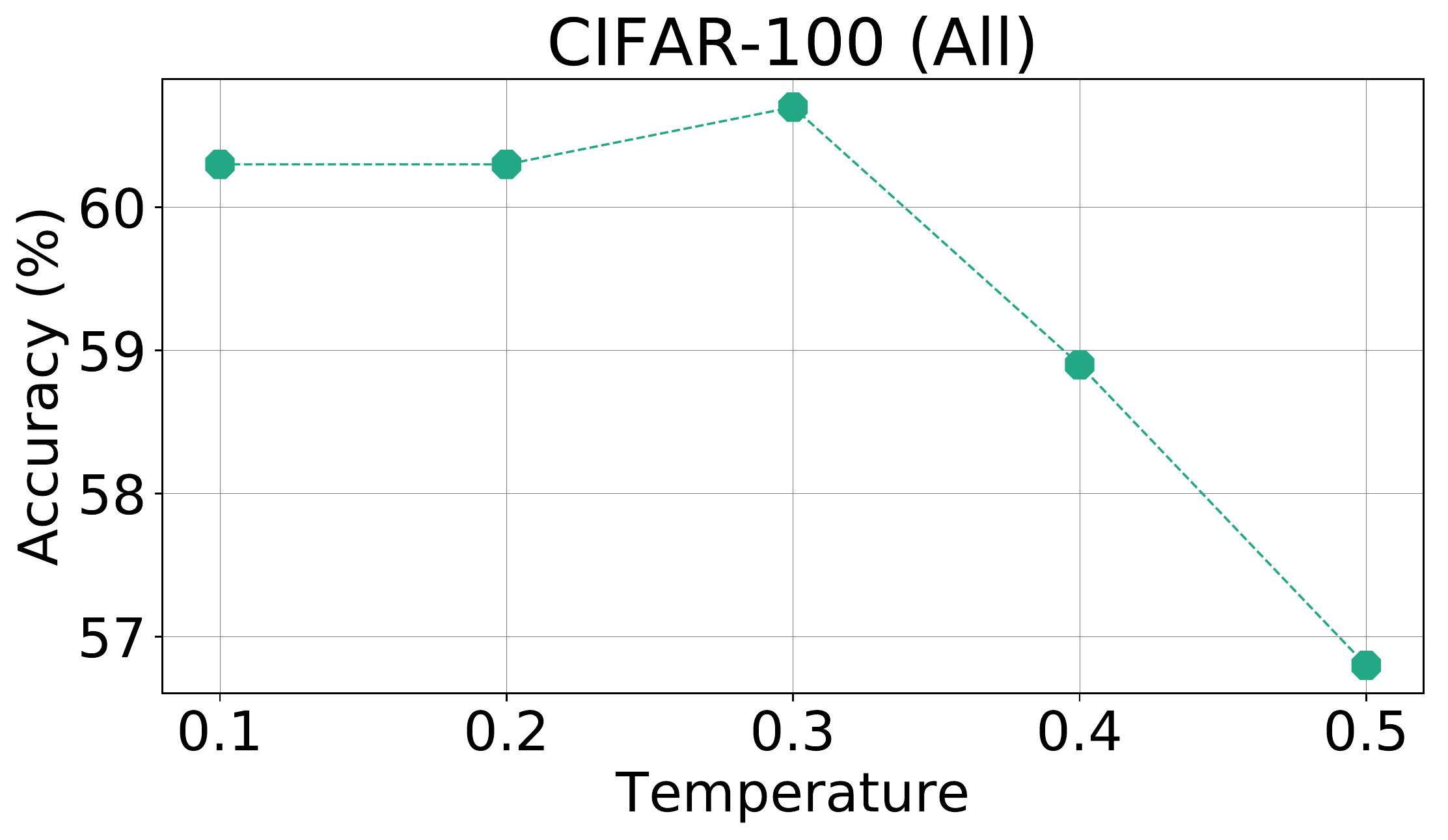}{\label{fig:temp_all}} }}
\vspace{-4mm}
    \caption{Accuracy on \textbf{CIFAR-100} dataset with different temperature values. These graphs suggest that our proposed method is not sensitive to change of temperature parameter over a large interval of temperature values.}
    \label{fig:temperature}
\end{figure*}

\section{Analysis of Temperature}
\label{sec:temp}
In this section, we discuss the effect of changing the temperature parameter of our proposed algorithm (Alg.~\ref{alg: alogirthm1}). To this end, we conduct a series of experiments by varying the temperature parameter on CIFAR-100 dataset with 10\% labeled data. The results are reported in Fig.~\ref{fig:temperature}. We notice that our proposed method is relatively stable over a large range of temperature values; even though we use a temperature of 0.1 in most of our experiments, results in Fig.~\ref{fig:temperature} suggest that temperature values of 0.2 and 0.3 also yield similar performances. However, the performance on seen/novel/all classes deteriorates for temperature values greater than 0.3. Overall these results demonstrate that the performance of our method is relatively stable to the choice of temperature hyperparameter. 

\begin{figure}[h]
\begin{center}
\includegraphics[width=\textwidth]{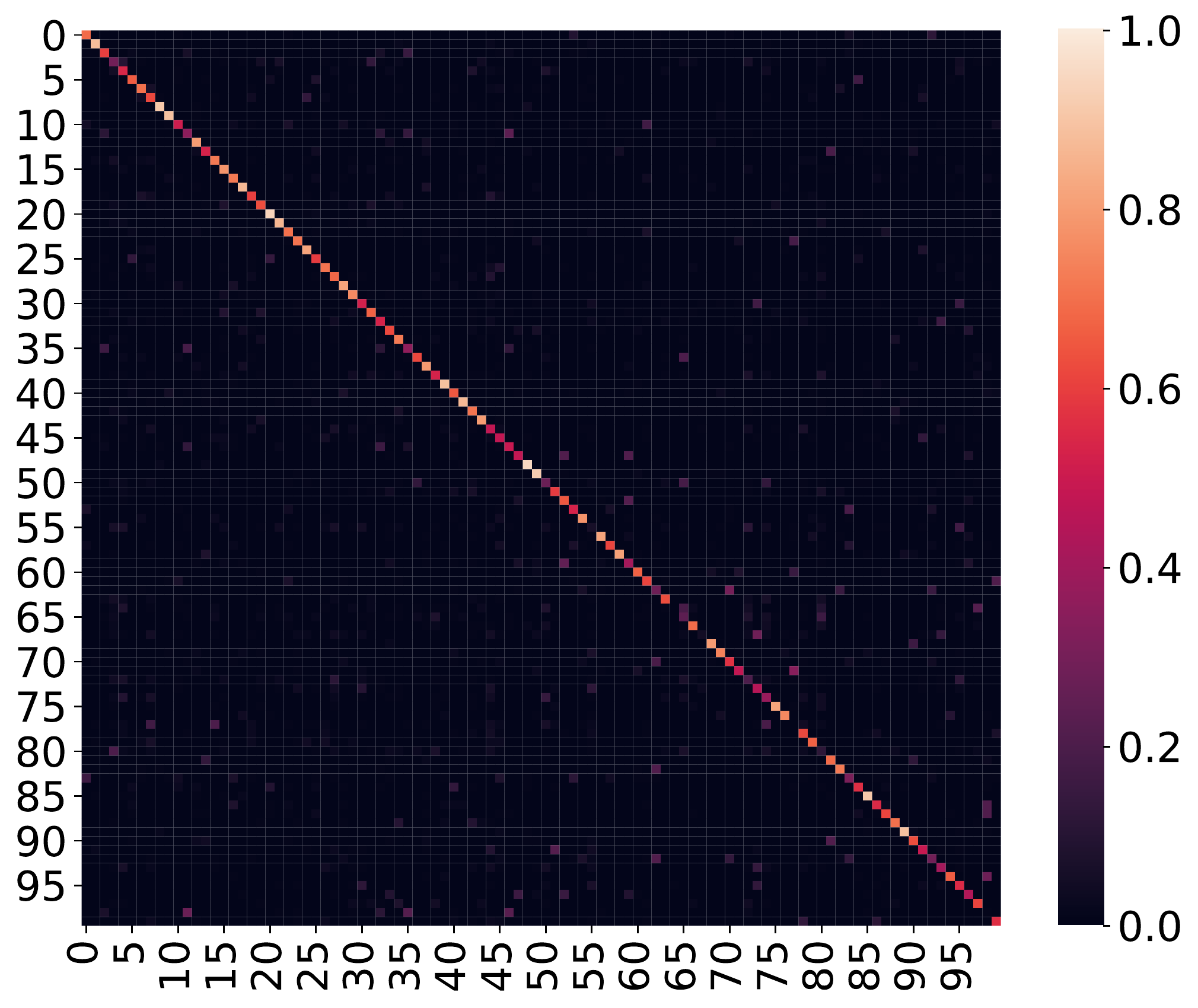}
\caption{Confusion matrix for test samples of \textbf{CIFAR-100} dataset. This matrix shows that our proposed method successfully recognises novel classes without confusing them with seen classes.}
\label{fig:confusion}
\end{center}
\end{figure}

\section{Confusing Novel Classes with Seen Classes}
\label{sec:conf_mat}
To perform an in-depth analysis of the performance of our proposed method on both seen and novel classes, we plot the confusion matrix for all the test samples of CIFAR-100 dataset (10\% labeeld data). In this analysis, we use Hungarian algorithm\cite{kuhn1955hungarian} to match the predictions for all classes to the ground-truth labels. We report the results in Fig.~\ref{fig:confusion}. These results provide evidence that our proposed algorithm can successfully recognise novel classes without confusing them with seen classes. Moreover, interestingly, our proposed method performs well even for a difficult problem, such as CIFAR-100, where it encounters a high number of novel classes.     

\clearpage
%
%
\bibliographystyle{eccv2022submission}
\bibliography{eccv2022submission}

\end{document}